\documentclass{article}

\usepackage{PRIMEarxiv}

\usepackage[utf8]{inputenc} 
\usepackage[T1]{fontenc}    
\usepackage{hyperref}       
\usepackage{url}            
\usepackage{booktabs}       
\usepackage{amsfonts}       
\usepackage{nicefrac}       
\usepackage{microtype}      
\usepackage{lipsum}
\usepackage{fancyhdr}       
\usepackage{graphicx}       
\graphicspath{{media/}}     
\usepackage{booktabs}       
\usepackage{amsfonts}       
\usepackage{nicefrac}       
\usepackage{microtype}      
\usepackage{xcolor}         
\usepackage{graphicx}

\usepackage{amsmath,amssymb,amsfonts,}
\usepackage{algorithmic}
\usepackage{algorithm}
\usepackage{amsmath}
\usepackage{subfigure}
\usepackage{multirow}
\usepackage{booktabs}
\usepackage{amssymb}
\usepackage{bbding}
\usepackage{pifont}
\usepackage{wasysym}
\usepackage{utfsym}
\usepackage{fontawesome}
\title{GrassNet: State Space Model Meets Graph Neural Network}

\author{
  Gongpei Zhao, Tao Wang, Congyan Lang, Yi Jin, Yidong Li \\
  School of Computer Science and Technology \\
  Beijing Jiaotong University \\
  Beijing\\
  \texttt{\{csgpzhao, twang\}@bjtu.edu.cn} \\
   \And
  Haibin Ling \\
  Department of Computer Science \\
  Stony Brook University \\
  Stony Brook\\
  \texttt{hling@cs.stonybrook.edu} \\
}

\begin{document}
\maketitle

\begin{abstract}
Designing spectral convolutional networks is a formidable task in graph learning. In traditional spectral graph neural networks (GNNs), polynomial-based methods are commonly used to design filters via the Laplacian matrix. In practical applications, however, these polynomial methods encounter inherent limitations, which primarily arise from the the low-order truncation of polynomial filters and the lack of overall modeling of the graph spectrum. This leads to poor performance of existing spectral approaches on real-world graph data, especially when the spectrum is highly concentrated or contains many numerically identical values, as they tend to apply the exact same modulation to signals with the same frequencies. To overcome these issues, in this paper, we propose \textbf{\textit{Gra}}ph \textbf{\textit{S}}tate \textbf{\textit{S}}pace  \textbf{\textit{Net}}work (GrassNet), a novel graph neural network with theoretical support that provides a simple yet effective scheme for designing and learning arbitrary graph spectral filters. In particular, our GrassNet introduces structured state space models (SSMs) to model the correlations of graph signals at different frequencies and derives a unique rectification for each frequency in the graph spectrum. To the best of our knowledge, our work is the first to employ SSMs for the design of GNN spectral filters, and it theoretically offers greater expressive power compared with polynomial filters. Extensive experiments on nine public benchmarks reveal that GrassNet achieves superior performance in real-world graph modeling tasks.
\end{abstract}


\section{Introduction}
As a class of neural networks (NNs) specifically designed to process and learn from graph data, graph neural networks (GNNs)~\cite{wu2020comprehensive,velivckovic2023everything} have gained significant popularity in addressing graph analytical challenges. They have demonstrated great success in various applications, including recommendation systems~\cite{wu2022graph}, drug discovery~\cite{xiong2021graph} and question answering~\cite{yasunaga2021qa}. Spatial-based and spectral-based GNNs are the two primary categories of graph neural networks. Different from spatial GNNs based on message passing and attention paradigms, spectral GNNs are built on graph spectral theory and graph filtering mechanism. They have attracted much research attention in recent few years, due to their solid theoretical foundation and  excellent practical performance.

Spectral GNNs map the graph signals to spectral domain through graph Fourier transform, design graph filters in spectral domain, and approximate graph convolutions with graph spectral filters. The design of spectral filters is crucial in the overall performance of GNN models, whose main idea is to modulate the frequency of a graph signal such that some of frequency components are amplified while others diminished. Recent studies suggest that most popular spectral GNNs operate as polynomial or wavelet graph spectral filters~\cite{defferrard2016convolutional,chien2021adaptive,levie2018cayleynets,bianchi2021graph,xu2018graph}. For example, GCN~\cite{kipf2016semi} uses a simplified first-order Chebyshev polynomial, which is proven to be a low-pass filter~\cite{balcilar2021analyzing}. ChebNet~\cite{defferrard2016convolutional} approximates the filtering operation with high-order Chebyshev polynomials. BernNet~\cite{he2021bernnet} adopts the Bernstein polynomial~\cite{phillips2003bernstein} to approximate spectral filters. The most recently proposed WaveNet~\cite{yang2024wavenet} utilizes the Haar wavelet bases~\cite{lepik2014haar} to approximate graph convolutions, where the filters are formulated as the sum of wavelet bases at various scales.

\begin{figure}[t!]
	\begin{center}
		\includegraphics[width=.70\linewidth]{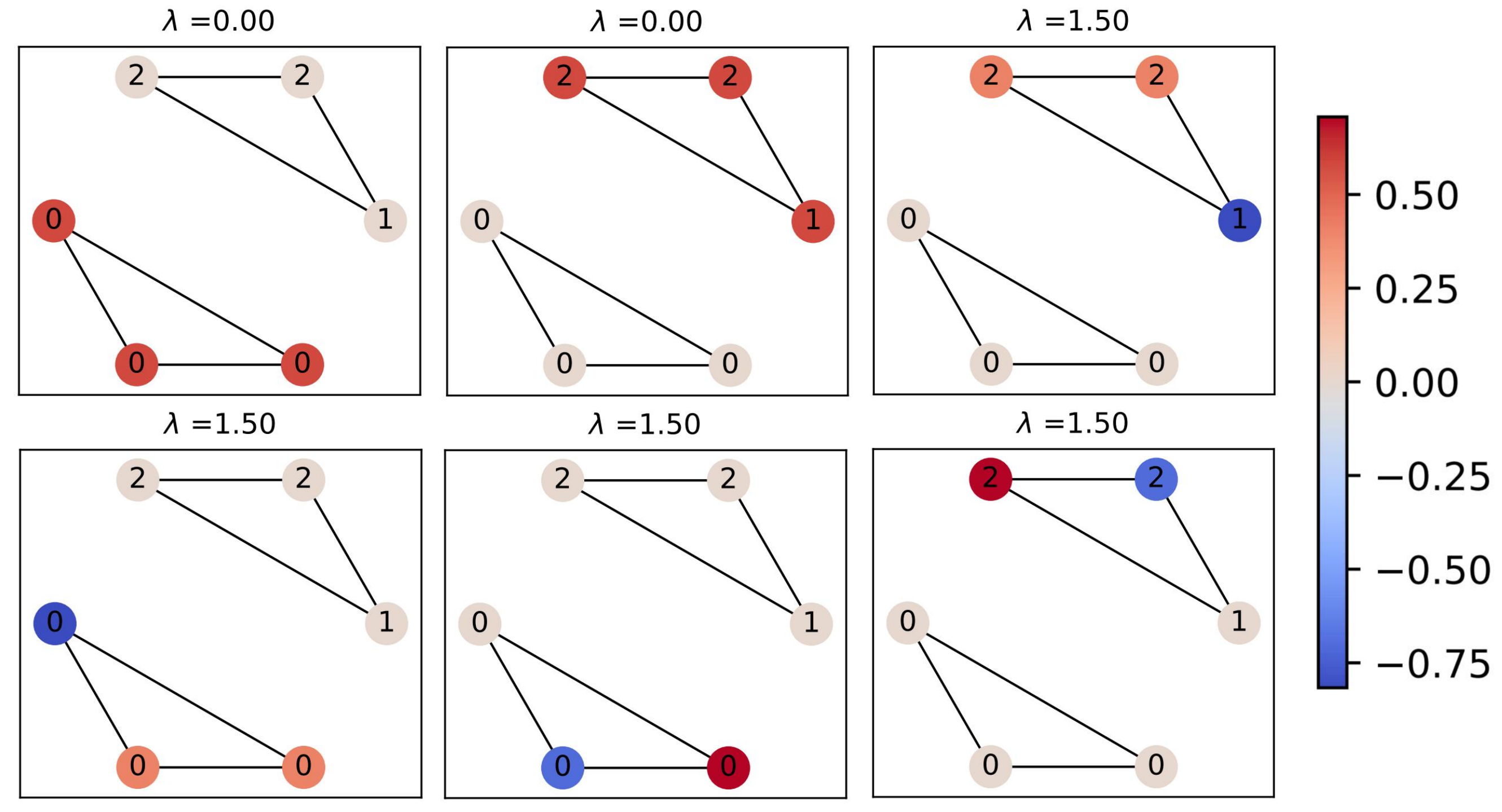}
	\end{center}
	\caption{Spectral decomposition of a toy graph. The values of the eigenvector corresponding to each eigenvalue are represented by the shading of the nodes, and the number on each node indicates its category.} 
	\label{fig:toy}
\end{figure}

Although remarkable results have been achieved by the aforementioned GNNs in various graph learning tasks, there remain some intriguing issues within spectral GNNs that are worth studying. \textbf{Firstly}, most existing methods focus on designing spectral filters based on predefined polynomial or wavelet bases, which requires extensive prior knowledge. This imposes strong inductive bias on GNN models and introduces numerous hyperparameters that need laborious tuning. For example, ChebNet with low-order polynomial bases often fails on heterophilic graphs where complex and unsmooth filters (\textit{e.g.}, band-rejection and comb filters) are required, whereas ChebNet with high-order polynomial bases suffers from overfitting on homophilic graphs that require simple and smooth filters. Strictly predefined inductive biases for spectral filters may impair the generality and generalization capability of models, and increase the difficulty of hyperparameter tuning. \textbf{Secondly}, existing spectral GNNs often overlook the global distribution of graph spectrum when designing filters, which to some extent may limit the modulation ability of filters. As shown in Fig.~\ref{fig:toy}, we consider a toy graph data with six nodes. Breaking down this graph by performing spectral decomposition on its normalized Laplacian matrix will yield two sets of numerically equal eigenvalues (\textit{i.e.}, 0 and 1.5). Each element in the set has the same frequency, but corresponds to different and mutually orthogonal graph signal bases. Existing spectral GNNs apply the same modulation to numerically equal frequencies, which is limiting. For example, in Fig.~\ref{fig:toy}, two frequencies with values equal to zero should not be subjected to the same modulation by an ideal filter, as the two connected components in the dataset exhibit distinctly different homophily. The graph signal in the top-left position should be preserved or enhanced, while the graph signal in the top-middle position should be attenuated. The same analysis can be applied to the other set of frequencies with a value equal to 1.5. Unfortunately, existing spectral filters based on high-order polynomial or wavelet fail to differentiate between spectral graph signal bases with the same frequency value. They typically apply the same modulation to numerically equal frequencies, which renders them ineffective in handling complex situations like the one shown in Fig.~\ref{fig:toy}.

To tackle these challenges, in this paper we propose \textbf{\textit{Gra}}ph \textbf{\textit{S}}tate \textbf{\textit{S}}pace  \textbf{\textit{Net}}work (GrassNet), a novel spectral graph neural network for semi-supervised node classification task. It does not impose strong inductive biases on spectral filters and can effectively handle complex situations where frequencies are numerically similar or equal. Specifically, our GrassNet models the entire sequence of the ordered spectrum of the graph data, allowing its spectral filter to fully capture the correlations between the elements within spectrum. This approach enables different enhancement/attenuation processes even for numerically identical frequencies. Given the recent advancements of structured state space sequence models (SSMs) and their widespread application in tasks within computer vision~\cite{zhuvision,yang2024vivim} and natural language processing~\cite{gu2023mamba,shan2024augmenting}, we design spectral filters based on the fundamental principles of SSMs to model graph spectrum sequences. Unlike existing spectral GNNs in which the filters are mappings defined in continuous space, our method transforms the spectral filtering issue into the modeling issue of discretized spectrum sequences. To the best of our knowledge, this is the first work to apply SSMs to the design of spectral GNN models.

In summary, the main contributions of this work lie in four significant folds:

\begin{itemize}
	\item \textbf{A new perspective of graph spectral filter:} We introduce a new discussion on polynomial and wavelet-based methods, point out their potential limitations, and propose using sequence models for spectral filtering.
	\item \textbf{Adaptation of SSMs for non-sequential graph data:} We design an effective way to extend SSMs to handle non-sequential graph data. Specifically, we design spectral filters based on the fundamental principles of SSMs to model graph spectrum sequences.
	\item \textbf{Innovative graph neural network design:} We propose GrassNet, a novel approach that represents a new type of spectral graph neural network pioneering the integration with SSMs.
	\item \textbf{Superior performance and efficiency:} We theoretically prove the advantages of GrassNet, and validate its effectiveness on nine benchmarks. The experimental results demonstrate the superiority of GrassNet against state-of-the-art spectral GNNs.
\end{itemize}


\section{Related Work}
\subsection{Spectral Graph Neural Networks}

Spectral GNNs operate by filtering graph signals in the spectral domain. ChebNet~\cite{defferrard2016convolutional} uses Chebyshev polynomials to approximate a spectral filter. APPNP~\cite{gasteiger2018predict} employs Personalized PageRank to derive the propagation matrix, functioning as a fixed filter. GPR-GNN~\cite{chien2021adaptive} introduces a learnable filtering model by learning the combination coefficients of the monomial basis. FAGCN~\cite{bo2021beyond} designs an adaptive self-gating mechanism to adaptively integrate different signals in the process of message passing. BernNet~\cite{he2021bernnet} leverages the Bernstein basis to learn a polynomial function which is more robust than monomial basis. ARMA~\cite{bianchi2021graph} learns a rational filter through the family of auto-regressive moving average filters. ChebNetII~\cite{he2022convolutional} seeks to improve upon the Chebyshev basis through interpolation to avoid the Runge phenomenon, making significant advancements over ChebNet. The recently proposed WaveNet~\cite{yang2024wavenet} uses Haar wavelet bases to approximate graph convolutions. In this approach, the filters are formulated as a sum of wavelet bases at various scales. However, the aforementioned methods predefine the mathematical form of the filters and cannot differentiate between numerically identical frequencies, thereby limiting the expressive power of GNNs.

\subsection{SSMs for Graph Learning}

State space models~\cite{guefficiently} (SSMs) are effective in modeling temporal and sequential data across domains like natural language processing~\cite{gu2023mamba,shan2024augmenting} and computer vision~\cite{zhuvision,yang2024vivim}. SSMs represent hidden states as functions of observed data over time, modeling latent dynamics and incorporating prior knowledge and constraints. This flexibility makes them suitable for capturing complex dependencies and structures in sequences. There has been some work that applies SSMs to graph learning tasks. Graph-Mamba~\cite{wang2024graph} is the first attempt to enhance long-range context modeling in graph networks by integrating a Mamba block with the input-dependent node selection mechanism. Similarly, GMNs~\cite{behrouz2024graph} adapts SSMs to graph-structured data which includes five processing steps, namely neighborhood tokenization, token ordering, architecture of bi-directional selective SSM encoder, local encoding and positional encoding. GSSC~\cite{huang2024can} introduces a principled extension of SSMs to graph data which preserves all advantages of SSMs by leveraging global permutation-equivariant set aggregation and factorizable graph kernels that rely on relative node distances as the convolution kernels. However, these methods all adapt SSMs to graph neural networks from a spatial perspective. In contrast, our work explores the feasibility of constructing graph filters through SSMs from a spectral perspective.

\section{Preliminaries}

\subsection{Problem Definition}

An attributed relational graph of $n$ nodes can be represented by $G=(\mathcal{V}, \mathcal{E}, \textbf{X})$, where $\mathcal{V}=\{v_{1}, v_{2},\cdots,v_{n}\}$ represents the set of \textit{n} nodes, and $\mathcal{E}=\{e_{ij}\}$ signifies the set of edges. $\textbf{X} =[\textbf{x}_{1}^{\top}; \textbf{x}_{2}^{\top};\cdots; \textbf{x}_{n}^{\top}] \in \mathbb{R}^{n \times d}$ is the attribute set for all nodes, with $\textbf{x}_{i}$ being the $d$-dimensional attribute vector for node $v_{i}$. The adjacency matrix $\mathcal{A} = \{a_{ij}\} \in \mathbb{R}^{n \times n}$ denotes the topological structure of graph \textit{G}, where $a_{ij}>0$ if there exists an edge $e_{ij}$ between nodes $v_{i}$ and $v_{j}$ and $a_{ij}=0$ otherwise. For semi-supervised node classification, the node set $\mathcal{V}$ can be split into a labeled node set $\mathcal{V}_{L} \subset \mathcal{V}$ with attributes $\textbf{X}_{L} \subset \textbf{X}$ and an unlabeled one $\mathcal{V}_{U}=\mathcal{V}/\mathcal{V}_{L}$ with attributes $\textbf{X}_{U}=\textbf{X}/\textbf{X}_{L}$.\footnote{For notation conciseness, we abuse the set notation and matrix notation interchangeably whenever appropriate. For example, $\textbf{X}$ represents both a set of $n$ attributes and a matrix.} We assume that each node belongs to exactly one class, and denote $\textbf{y}_{L}=\{y_{i}\}$ as the ground-truth labels of node set $\mathcal{V}_{L}$ where $y_{i}$ denotes the class label of node $v_{i} \in \mathcal{V}_{L}$.

The objective of semi-supervised node classification is to train a classifier using the graph and the known labels $\textbf{y}_{L}$, and apply this classifier to predict the labels for the unlabeled nodes $\textbf{v}_{U}$. Define a classifier 
$f_{\theta}$ as $(\tilde{\textbf{y}}_{L}, \tilde{\textbf{y}}_{U}) = f_{\theta}(\textbf{X}, \mathcal{A}, \textbf{y}_{L})$, where $\theta$ is the parameters of model. $\tilde{\textbf{y}}_{L}$ and $\tilde{\textbf{y}}_{U}$ are the predicted labels of nodes $\textbf{v}_{L}$ and $\textbf{v}_{U}$ respectively. Generally, the goal is to make the predicted labels $\tilde{\textbf{y}}_{L}$ align as closely as possible with the ground-truth labels $\textbf{y}_{L}$ in favor of: $\theta^{*}=\arg\min_{\theta}d(\tilde{\textbf{y}}_{L}, \textbf{y}_{L})=\arg\min_{\theta}d(f_{\theta}(\textbf{X}, \mathcal{A}, \textbf{y}_{L}), \textbf{y}_{L})$, where $d(\cdot,\cdot)$ denotes a measure of some type of distance between two sets of labels.

\subsection{Spectral Definition on Graph}

Given a graph whose adjacency matrix is represented as $\mathcal{A}$, the normalized adjacency matrix is defined as $\tilde{\mathcal{A}}=\textbf{D}^{-\frac{1}{2}}\mathcal{A}\textbf{D}^{-\frac{1}{2}}$, where $\textbf{D}$ is the degree matrix of $\mathcal{A}$. The normalized Laplacian matrix is denoted as $\textbf{L}=\textbf{I}-\tilde{\mathcal{A}}$, where $\textbf{I}$ denotes identity matrix, and the eigenvalues $\lambda_{i}$ lie within the interval $[0, 2]$. Let $\textbf{L}=\textbf{U}\textbf{$\Lambda$}\textbf{U}^{\top}$ denotes the eigendecomposition of normalized Laplacian matrix, where $\textbf{U}=[\textbf{u}_{1}, \textbf{u}_{2}, \cdots,\textbf{u}_{n}]$ is the eigenvector matrix, $\textbf{$ \Lambda $}=\texttt{diag}([\lambda_{1}, \lambda_{2}, \cdots, \lambda_{n}])$ is the diagonal eigenvalues matrix. For graph signal $\textbf{X}$, the graph Fourier transform is defined as $\textbf{X}'=\textbf{U}^{\top}\textbf{X}$, and the inverse transform is $\textbf{X}=\textbf{U}\textbf{X}'$. The spectral graph filter $g_{\theta}(\cdot)$ is a function with the frequency $\lambda_{i}$ as input which determines how the corresponding frequency component should be modulated. This process can be expressed as follows:
\begin{gather}
	\begin{aligned}
		\textbf{U}g_{\theta}(\Lambda)\textbf{U}^{\top}\textbf{X}&=\textbf{U} \ \texttt{diag}([g_{\theta}(\lambda_{1}), \cdots, g_{\theta}(\lambda_{n})]) \ \textbf{U}^{\top}\textbf{X} \\[4pt]
		&=(g_{\theta}(\lambda_{1})\textbf{u}_{1}\textbf{u}_{1}^{\top}+\cdots +g_{\theta}(\lambda_{n})\textbf{u}_{n}\textbf{u}_{n}^{\top})\textbf{X},
	\end{aligned}
\end{gather}
The filtering coefficients $[g_{\theta}(\lambda_{1}), g_{\theta}(\lambda_{2}), \cdots, g_{\theta}(\lambda_{n})]$ output by graph filter are the weights for the linear combination of the basis matrices $[\textbf{u}_{1}\textbf{u}_{1}^{\top}, \textbf{u}_{2}\textbf{u}_{2}^{\top}, \cdots, \textbf{u}_{n}\textbf{u}_{n}^{\top}]$.

\section{Proposed Method}

\subsection{Sequential Spectral Graph Filtering Paradigm}

\begin{figure}[t]
	\setlength{\abovecaptionskip}{-.0cm}  
	\begin{center}
		\subfigure[Cora]{
			\begin{minipage}[b]{0.28\textwidth}
				\centering
				\includegraphics[width=1\textwidth]{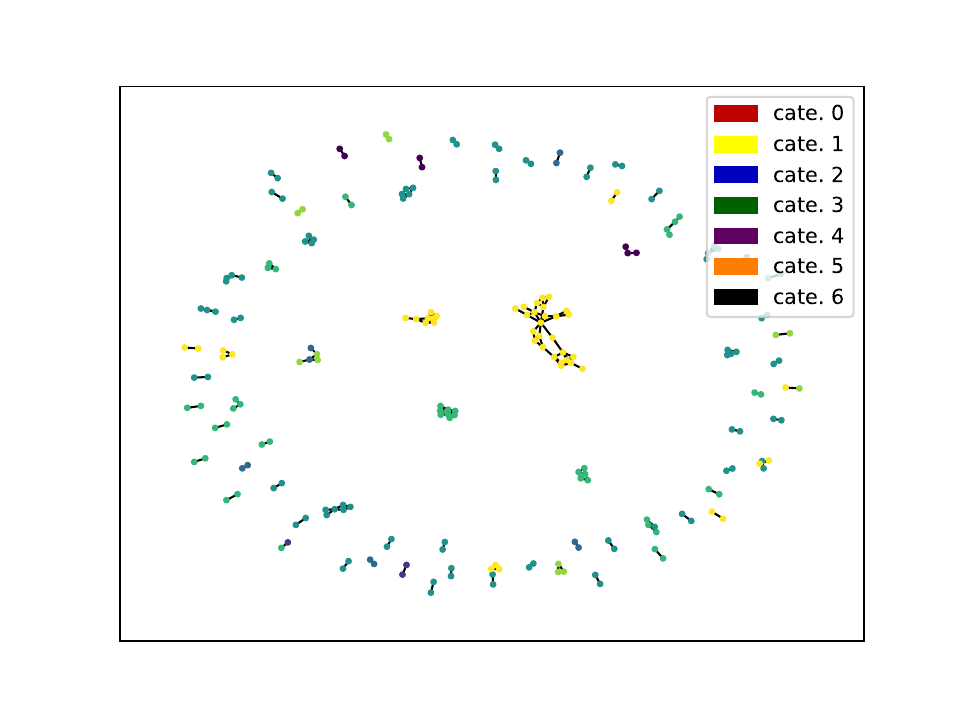}
			\end{minipage}
		}
		\subfigure[CiteSeer]{
			\begin{minipage}[b]{0.28\textwidth}
				\centering
				\includegraphics[width=1\textwidth]{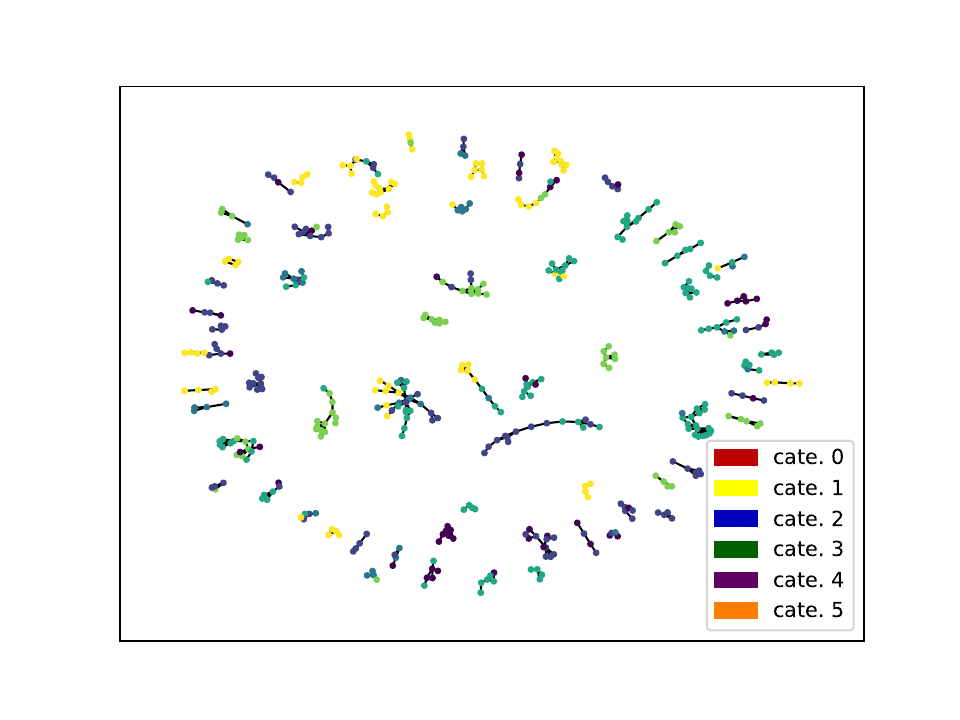}
			\end{minipage}
		}
		\subfigure[Computers]{
			\begin{minipage}[b]{0.28\textwidth}
				\centering
				\includegraphics[width=1\textwidth]{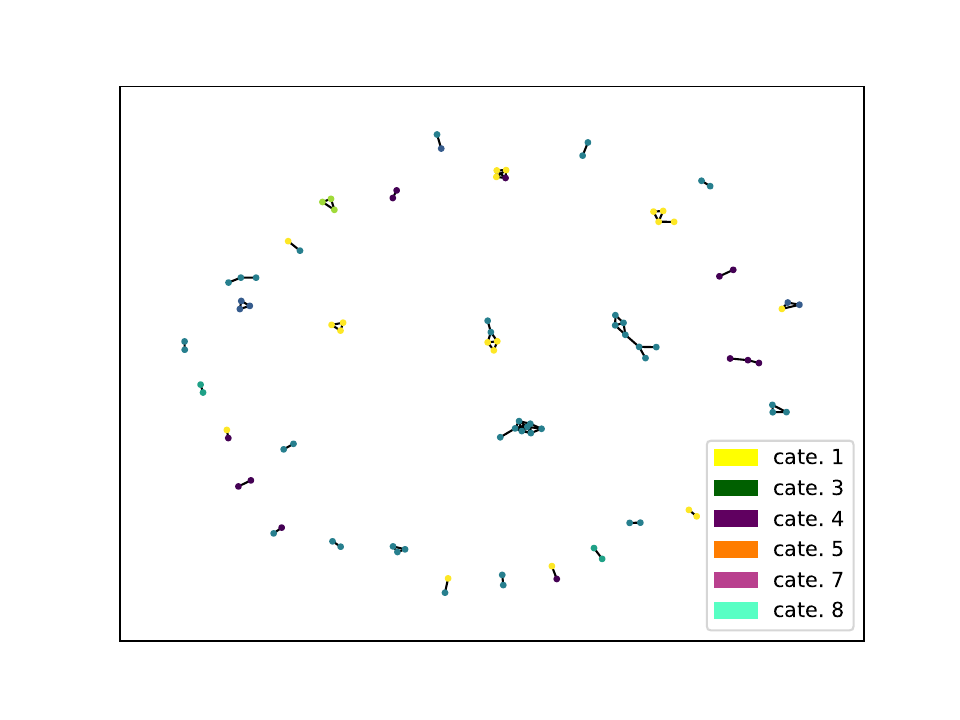}
			\end{minipage}
		}
	\end{center}
	\caption{Visualization of some connected components within three graph benchmarks.} 
	\label{fig:plot}
\end{figure}

Figure~\ref{fig:toy} illustrates the limitations of existing spectral graph filters in terms of modulation. This is mainly due to the fact that existing graph filters model each eigenvalue of Laplacian matrix independently, assuming all frequencies in the graph spectrum follow the independent and identically distributed (\textit{i.i.d.}) assumption without fully considering the potential correlations between them, \textit{e.g.}, the mutual orthogonality of signal bases corresponding to numerically identical frequencies. The overlooking is problematic because this issue commonly exists in real-world graph data. As shown in Fig.~\ref{fig:plot}, the connected components exhibit varying degrees of homophily. This implies that different filtering strategies should be applied for numerically close or even identical frequencies derived from different connected components. Specifically, for connected components with high homophily, low-frequency signals should be enhanced and high-frequency signals attenuated, whereas for heterophilic connected components, the opposite approach should be taken. Unfortunately, most spectral GNNs have limitations in effectively addressing this issue, potentially leading to performance bottlenecks. 

\begin{figure}[t]
	\setlength{\abovecaptionskip}{-.0cm}  
	\begin{center}
		\subfigure[Cora]{
			\begin{minipage}[b]{0.28\textwidth}
				\centering
				\includegraphics[width=1\textwidth]{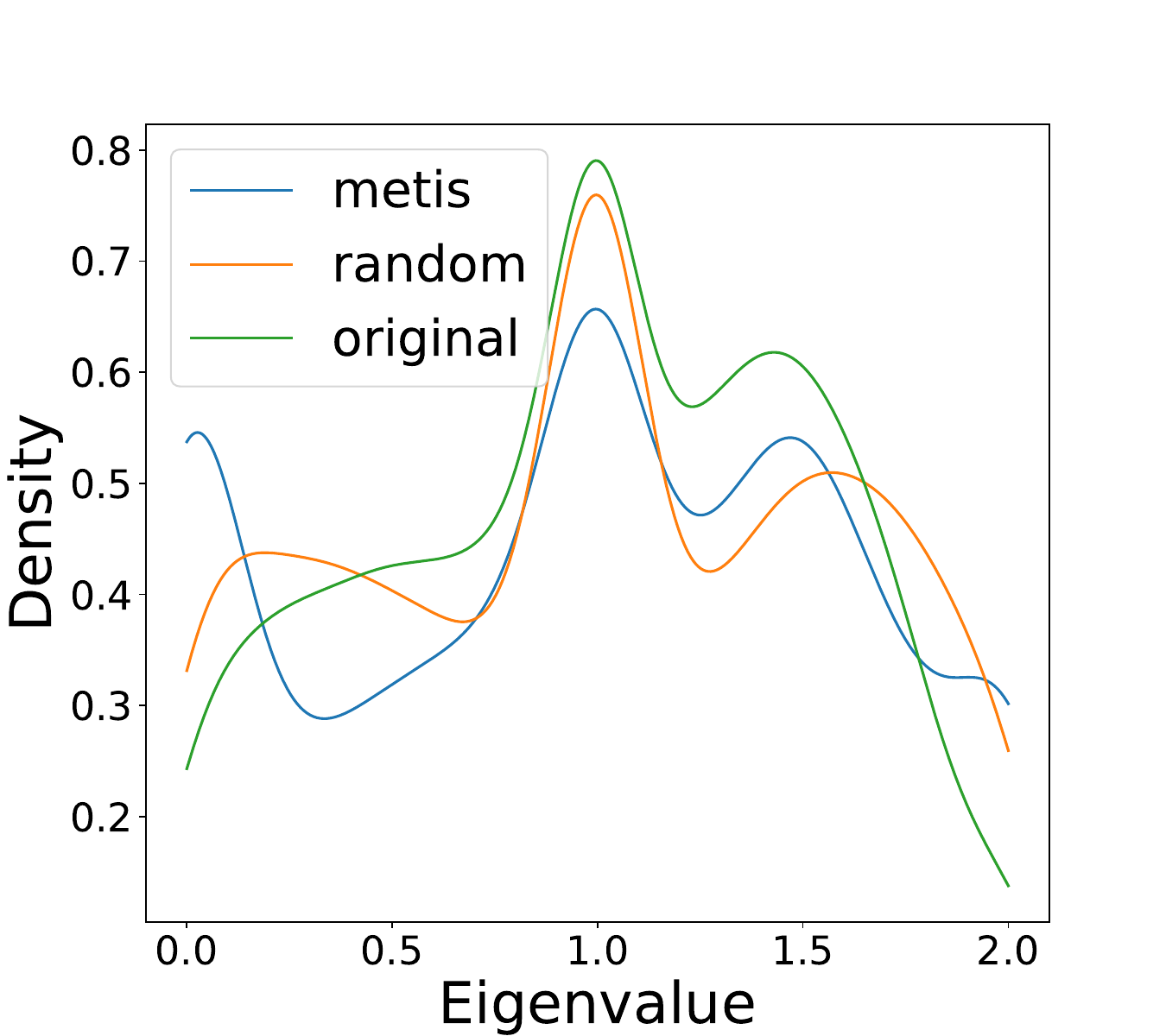}
			\end{minipage}
		}
		\subfigure[CiteSeer]{
			\begin{minipage}[b]{0.28\textwidth}
				\centering
				\includegraphics[width=1\textwidth]{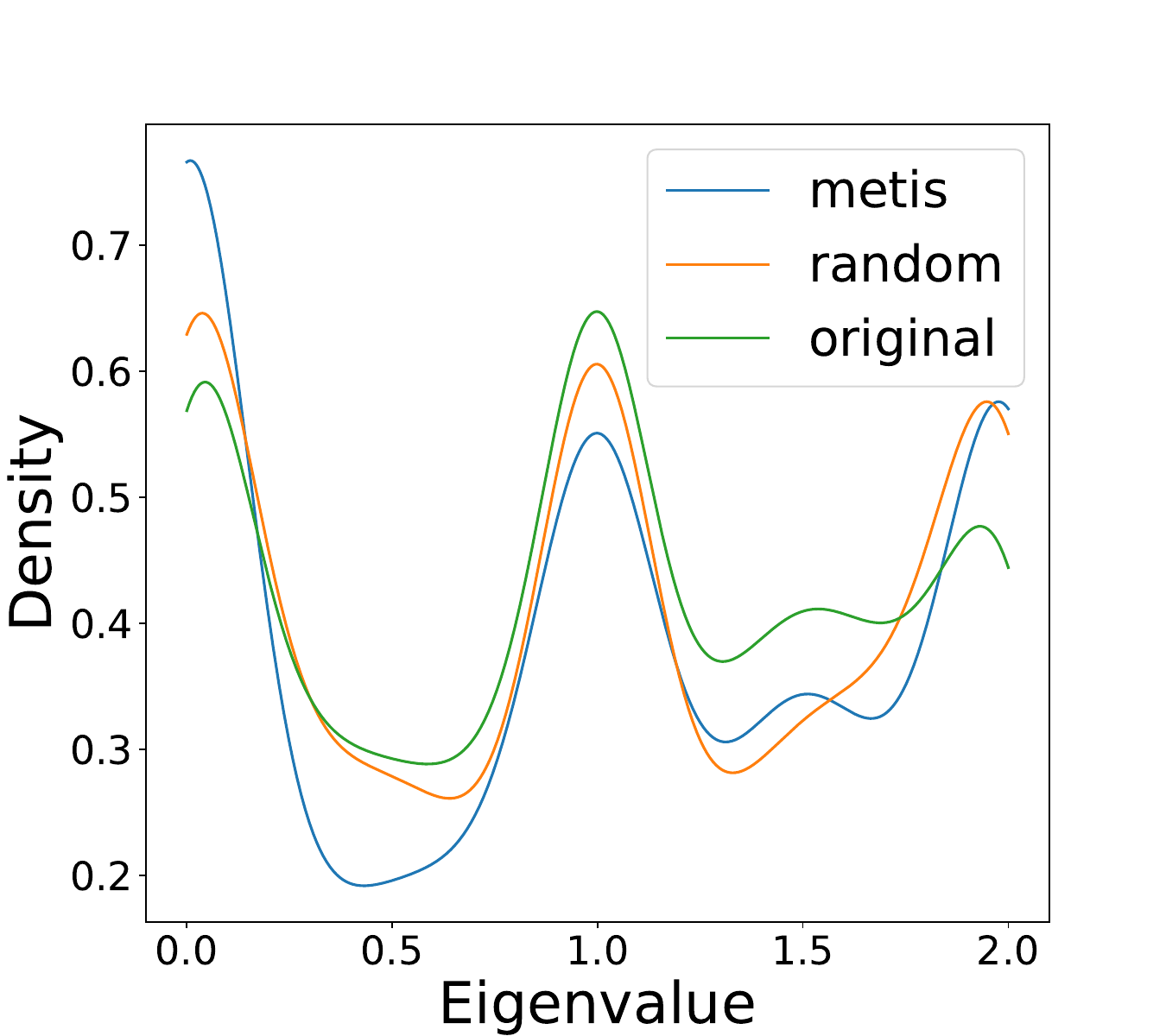}
			\end{minipage}
		}
		\subfigure[Photo]{
			\begin{minipage}[b]{0.28\textwidth}
				\centering
				\includegraphics[width=1\textwidth]{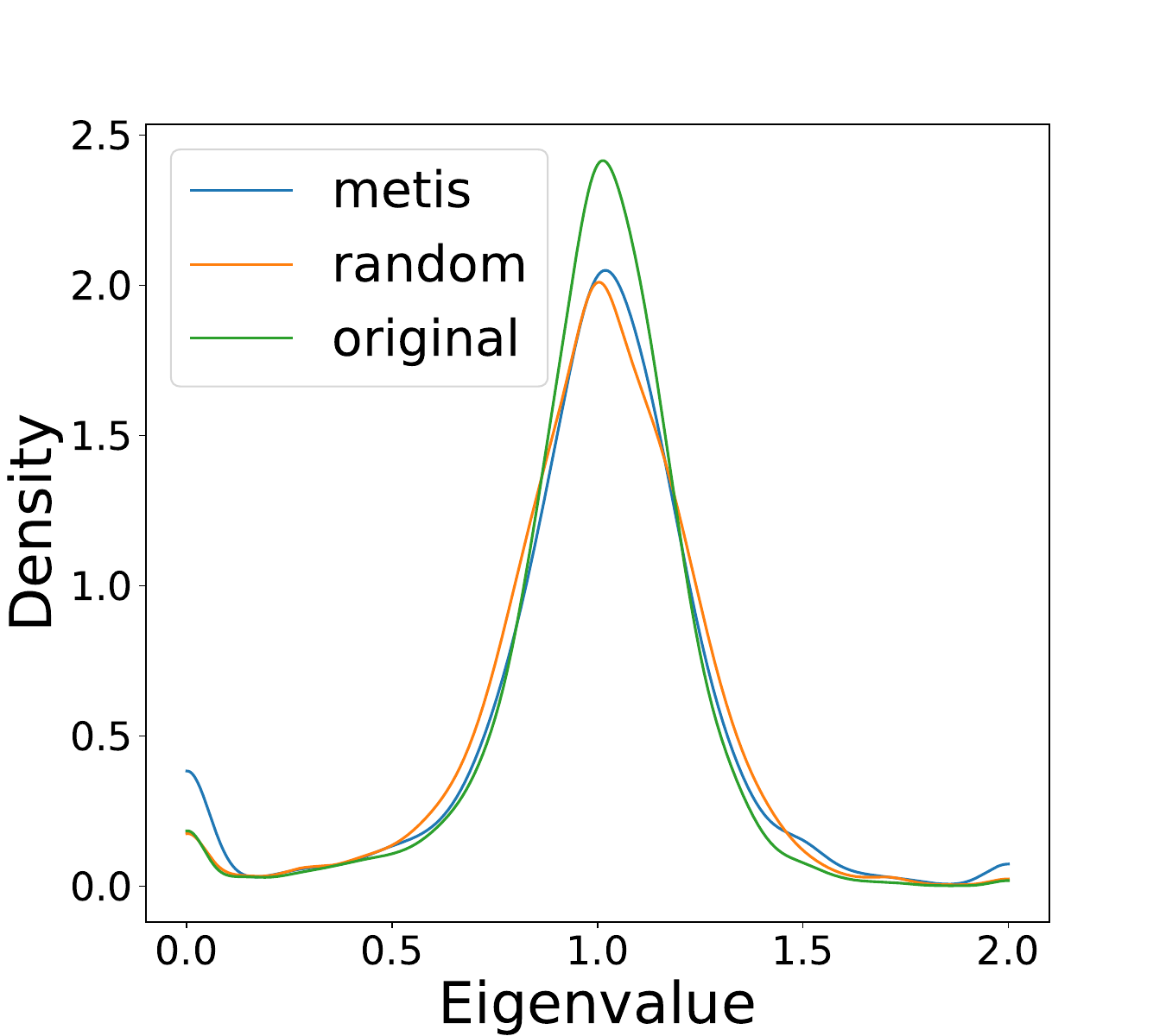}
			\end{minipage}
		}
		\subfigure[GCN]{
			\begin{minipage}[b]{0.45\textwidth}
				\centering
				\includegraphics[width=1\textwidth]{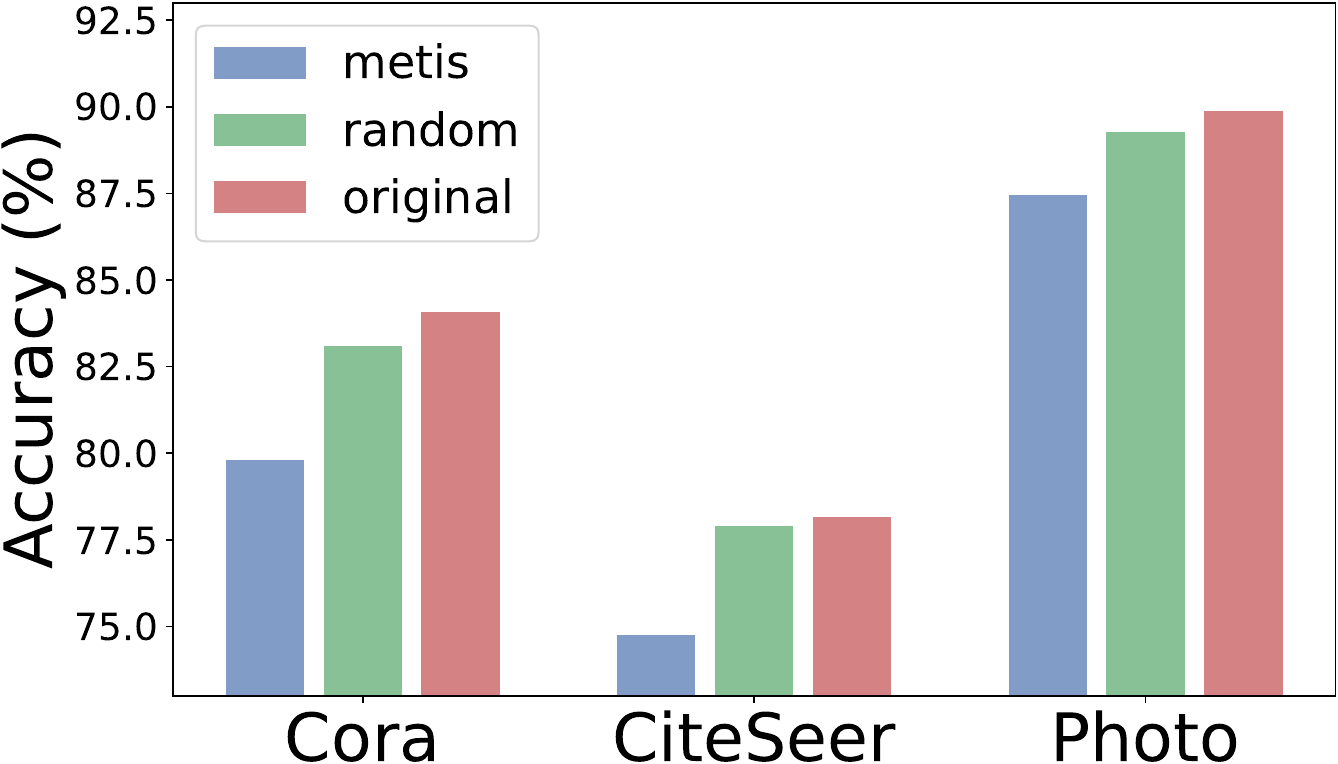}
			\end{minipage}
		}
		\subfigure[BernNet]{
			\begin{minipage}[b]{0.45\textwidth}
				\centering
				\includegraphics[width=1\textwidth]{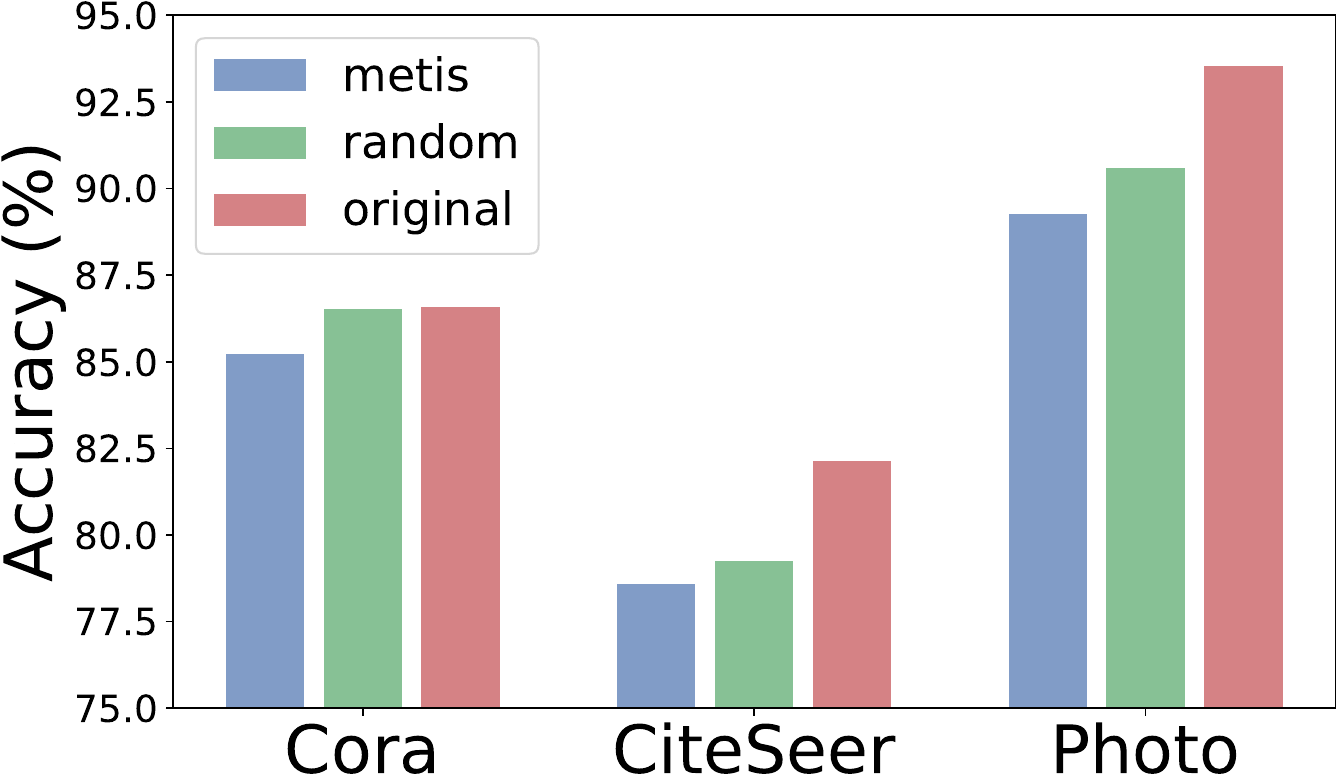}
			\end{minipage}
		}
	\end{center}
	\caption{(a-c): The kernel density estimation (KDE) on Cora, CiteSeer and Photo. (d-e): The classification accuracy of GCN and BernNet on the original graph topology and two modified graph topologies.} 
	\label{fig:attack}
\end{figure}

As shown in Fig.~\ref{fig:attack}, we adjust the original topology of three graph datasets. We perform Metis partitioning~\cite{karypis1997metis} to remove some edges from original graph. The number of removed edges is positively correlated with the original graph structure. This process tends to create more connected components, resulting in more numerically identical frequencies. We also randomly remove the same number of edges for comparison. It is shown in Fig.~\ref{fig:attack} (a-c) both methods increase the frequency density in the low-frequency and high-frequency regions, however, the effect is more pronounced with Metis, as it is a graph partitioning algorithm. The experimental results of well-known GCN and BernNet show existing spectral GNNs are not well-equipped to handle this issue and thus perform significantly worse than original and random trials.

To overcome these limitations, we advocate for changing the design paradigm of existing filters. In earlier studies the filter function $g_{\theta}(\cdot)$ is usually represented using various polynomial or wavelet basis as:
\begin{gather}\label{eq:poly}
	g_{\theta}(\lambda_{i})=\sum_{k=0}^{K}b_{k}\cdot \phi_{k}(\lambda_{i}),
\end{gather} 
where $\phi_{k}(\cdot)$ is the $k$-th order term of a predefined polynomial (\textit{e.g.}, power, Chebyshev or Bernstein polynomial) or wavelet at translation position $k$, $b_{k}$ the weight of the $k$-th term. Differently, we propose to capture the correlations between the frequencies, which could be formulated as:
\begin{gather}\label{eq:poly1}
	g_{\theta}(\lambda_{i})=\varphi(\psi(\lambda_{i}),[\psi(\lambda_{1}), \cdots, \psi(\lambda_{i}), \cdots, \psi(\lambda_{n})]),
\end{gather} 
where $\psi(\cdot)$ maps the frequency to a latent space, $\varphi$ takes the representation of $\lambda_{i}$ in the latent space and the sequence of representations of the ordered spectrum in the latent space as input, and outputs the filtering result of $\lambda_{i}$.

Compared with Eq.~\ref{eq:poly}, the proposed Eq.~\ref{eq:poly1} treats the ordered spectrum of graph data as a sequence with discrete values, allowing the spectral filter to fully capture the correlations between elements in the spectrum. This approach enables different enhancement/attenuation processes even for numerically identical frequencies. The function $\psi(\cdot)$ is flexible, with one option being to use the corresponding eigenvector of $\lambda_{i}$. Nevertheless, the computation results of eigenvectors are not deterministic, and the dimensions of eigenvectors increase with the size of the graph data, making it impractical to implement on large graphs. Existing approaches, such as Chebyshev or Bernstein polynomial functions, are also feasible, but they introduce manually predefined inductive biases to the model. We implement  $\psi(\cdot)$ using simple fully connected layers (FC), for its universal approximation property. 

For $\varphi(\cdot)$, we need to choose a sequence model. The popular Transformer~\cite{vaswani2017attention}, due to its high spatial complexity, is not ideal for inference on large-scale graphs, while recurrent neural networks (RNNs)~\cite{sundermeyer2012lstm,dey2017gate} incur significant time cost because of their strict sequential nature. As SSMs achieve partial computational parallelism and have low spatial complexity, we integrate SSMs into our spectral filters in Eq.~\ref{eq:poly1}. Compared with the paradigm in Eq.~\ref{eq:poly}, our approach incorporates the process of sequential modeling of frequencies, thereby theoretically offering greater expressive power than existing methods.
\begin{figure*}[t]
	\begin{center}
		\includegraphics[width=1\textwidth]{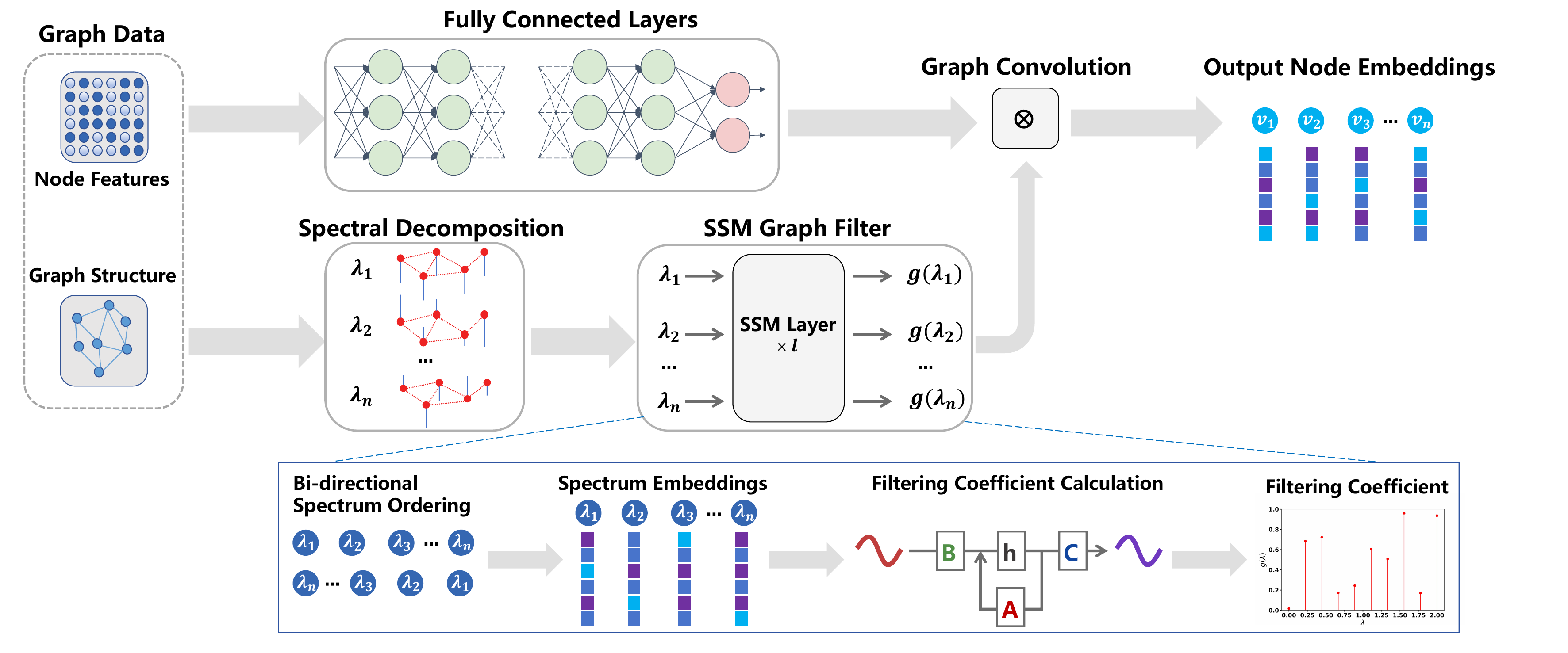}
	\end{center}
	\caption{Overall framework of the proposed GrassNet. The SSM graph filter is composed of cascaded SSM layers, which takes the ordered graph spectrum as input and outputs a filtering coefficient for each frequency within the spectrum. With the learned filtering coefficients, graph convolution is then applied to the latent embeddings of nodes, derived from fully connected layers.} 
	\label{fig:framework}
\end{figure*}

\subsection{Spectral Graph Filters based on SSMs}\label{sec:ssm}

SSMs are recurrent models and require ordered input. To this end we derive the sequence $[\psi(\lambda_{1}), \cdots, \psi(\lambda_{i}), \cdots, \psi(\lambda_{n})]$ by ordering according to the value of $\lambda_{i}$. Let $\textbf{H}\in \mathbb{R}^{n\times d_\text{in}}$ be the input sequence in which the $i$-th row is $\psi(\lambda_{i})\in \mathbb{R}^{d_\text{in}}$, we have:
\begin{gather}
	\textbf{B}=\textbf{H}\textbf{W}_{\textbf{B}}, \ \  \textbf{C}=\textbf{H}\textbf{W}_{\textbf{C}}, \ \ \textbf{$\Delta$}=\texttt{softplus}(\textbf{H}\textbf{W}_{\Delta}), \label{eq:selection}\\[4pt]
	\bar{\textbf{A}}, \bar{\textbf{B}}=\texttt{discretize}(\Delta, \textbf{A}, \textbf{B}), \label{eq:discretize}\\[5pt]
	\textbf{S}=\texttt{SSM}_{\bar{\textbf{A}}, \bar{\textbf{B}}, \textbf{C}}(\textbf{H}), \ \ \textbf{s}=\textbf{S}\textbf{w}_{\textbf{O}},\label{eq:out}
\end{gather}
where $\textbf{W}_{\textbf{B}}$, $\textbf{W}_{\textbf{C}}\in \mathbb{R}^{d_\text{in}\times d_\text{mid}}$, $\textbf{W}_{\Delta}\in \mathbb{R}^{d_\text{in}\times d_\text{in}}$, $\textbf{w}_{\textbf{O}}\in \mathbb{R}^{d_\text{in}}$ are learnable parameters, and the structure of $\textbf{A}\in \mathbb{R}^{d_\text{in}\times d_\text{mid}}$ follows \cite{gu2023mamba}, which represents $d_\text{in}$ diagonal $d_\text{mid}\times d_\text{mid}$ matrices using a learnable $d_\text{in}\times d_\text{mid}$ matrix. Equation~\ref{eq:selection} formulates the selection mechanism applied in S6~\cite{gu2023mamba}. This process is achieved by making the SSM parameters $\textbf{B}$, $\textbf{C}$ and $\Delta$ functions of the input $\textbf{H}$. This allows the model to adaptively select relevant information from the context. The discretization process $\texttt{discretize}(\cdot)$ in Eq.~\ref{eq:discretize} is formulated as $\bar{\textbf{A}}=\exp(\Delta\textbf{A})$, $\bar{\textbf{B}}=(\Delta\textbf{A})^{-1}(\exp(\Delta\textbf{A})-\textbf{I})\Delta\textbf{B}$. $\texttt{SSM}(\cdot)$ in Eq.~\ref{eq:out} is the state space model formulated as:
\begin{gather}
	\textbf{h}_{i}=\bar{\textbf{A}}\textbf{h}_{i-1}+\bar{\textbf{B}}\psi(\lambda_{i}), \ \ \textbf{s}_{i}=\textbf{C}\textbf{h}_{i},
\end{gather}
where $\textbf{h}_{i}\in \mathbb{R}^{d_\text{mid}}$ is the latent state, $\textbf{s}_{i}\in \mathbb{R}^{d_\text{in}}$ the output of SSM, which is the $i$-th row of $\textbf{S}\in \mathbb{R}^{n\times d_\text{in}}$. We use a fully-connected layer parameterized as $\textbf{w}_{\textbf{O}}$ to derive the filtering coefficients $\textbf{s}\in \mathbb{R}^{n}$, and rescale to constrain the values of the elements in $\textbf{s}$ as $\hat{\textbf{s}}=\gamma\cdot\textbf{s}/|\textbf{s}|_{max}$, where $|\cdot|_{max}$ denotes the maximum absolute value within $\textbf{s}$, and $\gamma$ is the scaling factor. Equations~\ref{eq:selection}, \ref{eq:discretize} and \ref{eq:out} implement the sequential spectral graph filtering paradigm in Eq.~\ref{eq:poly1}. For SSMs, a GPU-friendly implementation~\cite{gu2023mamba} is developed for efficient computation of the selection mechanism, which drastically reduces memory I/Os and eliminates the need to store intermediate states.

\subsection{Graph State Space Network (GrassNet)}

The overall framework of proposed GrassNet is illustrated in Fig.~\ref{fig:framework}. GrassNet incorporates spectral graph filters based on SSMs into its framework, which is moduled as SSM graph filter. This filter derives the spectrum embedding for each frequency according to $\psi(\cdot)$ defined in Eq.~\ref{eq:poly1}, and then scans the input sequence in two different directions, enabling each frequency to receive information from all higher or lower frequencies. The bi-directional design enhances the robustness of filter to permutations and enables it to learn equivariant functions on the input whenever needed. The SSM graph filter consists of $l$ stacked SSM layers and outputs the filtering coefficient for each frequency. With the filtering coefficients and latent representations of nodes obtained by a fully connected layer, the graph convolution operation transforms these latent embeddings from spatial domain to spectral domain. In spectral domain, the graph signals are adjusted based on the filtering coefficients, and then transformed back to the spatial domain. The convolution process is formulated as follows:
\begin{align}
	\hat{\textbf{X}}&= \texttt{FC}(\textbf{X}), \label{eq:fc} \\[2pt]
	\tilde{\textbf{X}}&=\textbf{U} \ \texttt{diag}([g(\lambda_{1}), \cdots, g(\lambda_{n})]) \ \textbf{U}^{\top}\hat{\textbf{X}}, \label{eq:conv}
\end{align}
where $\texttt{FC}(\cdot)$ denotes the fully connected layers, $\tilde{\textbf{X}}$ and $\hat{\textbf{X}}$ are latent embeddings and output node embeddings, respectively. The output node embeddings could be used for downstream tasks. For example, a simple linear classifier could be added afterward to perform end-to-end node classification.

Our method comprises three main steps: spectral decomposition, filter learning, and graph convolution. Since the time complexity of the spectral decomposition is $\mathcal{O}(n^{3})$, we perform the decomposition as a preprocessing step. The time complexity of filter learning largely depends on SSM graph filter, which is $\mathcal{O}(n)$, while the time complexity of graph convolution is $\mathcal{O}(n^{2})$. However, all the three steps can be parallelized to reduce time consumption.

Compared with other potentially accessible methods for sequential spectral graph filtering, our method  show significant superiority. The popular attention mechanism requires $\mathcal{O}(n^{2})$ space complexity. This limits the scale of graph data. RNNs have a time and space complexity of $\mathcal{O}(n)$, which is the same as our SSM graph filter. However, the parallel associative scan strategy and hardware aware algorithm within SSMs contributes to the parallel computing, resulting in logarithmic time complexity in the number of sub-intervals, which could be theoretically smaller than $\mathcal{O}(n)$.

As the spectral filtering of GrassNet is global, which is different from spatial methods that rely on local message passing, we perform graph convolution only once in the entire process, as formulated in Eq.~\ref{eq:conv}. In fact, the operation of Eq.~\ref{eq:conv} could be regarded as global message passing. The operations in Eqs.~\ref{eq:fc} and \ref{eq:conv} disentangle message passing from feature transformation, which is beneficial for alleviating model degradation and oversmoothing~\cite{zhang2022model}.

\begin{table*}[t!]
	\centering
	\setlength\tabcolsep{1.4pt}
	\renewcommand{\arraystretch}{1.5} 
	\caption{Results on datasets: mean accuracy (\%) $\pm$ 95\% confidence interval. The best and the second results on each dataset are indicated with \textbf{bold} and \underline{underline}, respectively.}\label{tab:comparison}
	\begin{tabular}{cccccccccc}
		\hline
		& Cora       & CiteSeer   & PubMed      & Photo      & Chameleon  & Squirrel  & Actor      & Texas      & Cornell    \\ \hline
		MLP        & 76.96$\pm$0.95 & 76.58$\pm$0.88 & 85.94$\pm$0.22  & 84.72$\pm$0.34 & 46.85$\pm$1.51 & 31.03$\pm$1.18 &40.19$\pm$0.56 & 91.45$\pm$1.14 & 90.82$\pm$1.63 \\
		GCN        & 84.07$\pm$1.01 & 78.18$\pm$0.67 & 86.74$\pm$0.27  & 89.88$\pm$0.73 &62.14$\pm$2.21 & 46.78$\pm$0.87  &33.23$\pm$1.16& 77.38$\pm$3.28 & 65.90$\pm$4.43 \\
		ChebNet   & 86.67$\pm$0.82 & 79.11$\pm$0.75 & 87.95$\pm$0.28  & 93.77$\pm$0.32 & 59.28$\pm$1.25 & 40.55$\pm$0.42  &37.61$\pm$0.89& 86.22$\pm$2.45 & 83.93$\pm$2.13 \\
		GAT        & 88.03$\pm$0.79 & 80.52$\pm$0.71 & 87.04$\pm$0.24  & 90.94$\pm$0.68 & 63.13$\pm$1.93 & 44.49$\pm$0.88  &33.93$\pm$2.47& 80.82$\pm$2.13 & 78.21$\pm$2.95 \\
		APPNP      & 88.14$\pm$0.73 & 80.47$\pm$0.74 & 88.12$\pm$0.31  & 88.51$\pm$0.31 & 51.84$\pm$1.82 & 34.71$\pm$0.57  &39.66$\pm$0.55& 90.98$\pm$1.64 & 91.81$\pm$1.96 \\
		GPR-GNN    & \underline{88.57}$\pm$0.69 & 80.12$\pm$0.83 & 88.46$\pm$0.33  & 93.85$\pm$0.28 & 67.28$\pm$1.09 & 50.15$\pm$1.92  &39.92$\pm$0.67& 92.95$\pm$1.31 & 91.37$\pm$1.81 \\
		SplineCNN  & 86.90$\pm$0.47 & 78.43$\pm$0.64 & 86.73$\pm$0.39  & 93.66$\pm$0.41 & 51.90$\pm$1.35 & 37.75$\pm$0.92  &35.11$\pm$0.82& 88.03$\pm$2.79        & 82.30$\pm$3.44        \\
		BernNet    & 88.52$\pm$0.95 & 80.09$\pm$0.79 & 88.48$\pm$0.41  & 93.63$\pm$0.35 & 68.29$\pm$1.58 & 51.35$\pm$0.73 &\textbf{41.79}$\pm$1.01& 93.12$\pm$0.65 & 92.13$\pm$1.64 \\
		ChebNetII & 88.10$\pm$0.93 & \underline{80.53}$\pm$0.79 & 88.93$\pm$0.29  & 94.06$\pm$0.37 & 71.37$\pm$1.01 & 57.72$\pm$0.59  &\underline{41.75}$\pm$1.07& \underline{93.28}$\pm$1.47 & \textbf{92.30}$\pm$1.48 \\
		WaveNet    & 88.46$\pm$0.51 & 80.22$\pm$0.71 & \textbf{90.26}$\pm$0.22  & \underline{94.42}$\pm$0.31 & \underline{72.60}$\pm$1.18 & \textbf{62.97}$\pm$1.18  &39.69$\pm$1.54& 91.48$\pm$1.97        & 81.15$\pm$3.12        \\ \hline
		ours       & \textbf{88.59}$\pm$0.69 & \textbf{81.51}$\pm$0.67 & \underline{89.13}$\pm$0.65       & \textbf{94.49}$\pm$0.27 & \textbf{72.71}$\pm$0.99 & \underline{61.60}$\pm$0.54  &40.56$\pm$0.68& \textbf{93.44}$\pm$2.13 & \underline{92.30}$\pm$2.13
		\\ \hline
	\end{tabular}
\end{table*}

\section{Experiments}

\subsection{Comparison with Baseline Algorithms}\label{app:dataset}

We evaluate our method on nine benchmark datasets across domains: citation networks (Cora, CiteSeer, PubMed)~\cite{sen2008collective,yang2016revisiting}, Amazon co-purchase graph (Photo)~\cite{mcauley2015image}, Wikipedia graphs (Chameleon, Squirrel)~\cite{rozemberczki2021multi}, actor co-occurrence graph (Actor)~\cite{pei2019geom} and webpage graphs from WebKB (Texas,  Cornell)~\cite{pei2019geom}. The detailed dataset statistics are summarized in Appx. B. 

We conduct semi-supervised node classification task with each baseline model, in which the node set is randomly divided into the train/validation/test set with 60\%/20\%/20\%. For fairness, we generate ten random splits using different seeds and evaluate all approaches on these identical splits, reporting the average performance for each method. The baseline models includes MLP~\cite{taud2018multilayer}, GCN~\cite{kipf2016semi}, GAT~\cite{velivckovic2018graph}, ChebNet~\cite{defferrard2016convolutional}, SplineCNN~\cite{fey2018splinecnn}, APPNP~\cite{gasteiger2018predict}, GPR-GNN~\cite{chien2021adaptive} and recently proposed BernNet~\cite{he2021bernnet}, ChebNetII~\cite{he2022convolutional} and WaveNet~\cite{yang2024wavenet}. All the GNN baseline models we compare are either spectral GNN models or GNN models that can be interpreted from a spectral perspective. The evaluation metric used is accuracy (acc), presented with a 95\% confidence interval. The hyperparameters for each baseline algorithms are set according to their official researches. The detailed experimental settings for our method are summarized in Appx. C.

The comparative analysis of various algorithms on benchmark datasets is summarized in Tab.~\ref{tab:comparison}. Our method demonstrates results comparable to state-of-the-art spectral GNNs, achieving top-2 performance on all datasets except for Actor, where it still ranks close to the top. Although our method underperforms BernNet and ChebNetII on Actor, the accuracy gap is small, and it significantly outperforms the other eight methods. These results highlight the strong performance of the proposed GrassNet in handling various real-world graphs and confirm that SSMs are effective spectral filters for complex graph spectral signals.

\begin{figure}[t]
	\setlength{\abovecaptionskip}{-.0cm}  
	\begin{center}
		\subfigure[Cora]{
			\begin{minipage}[b]{0.28\textwidth}
				\centering
				\includegraphics[width=.85\textwidth]{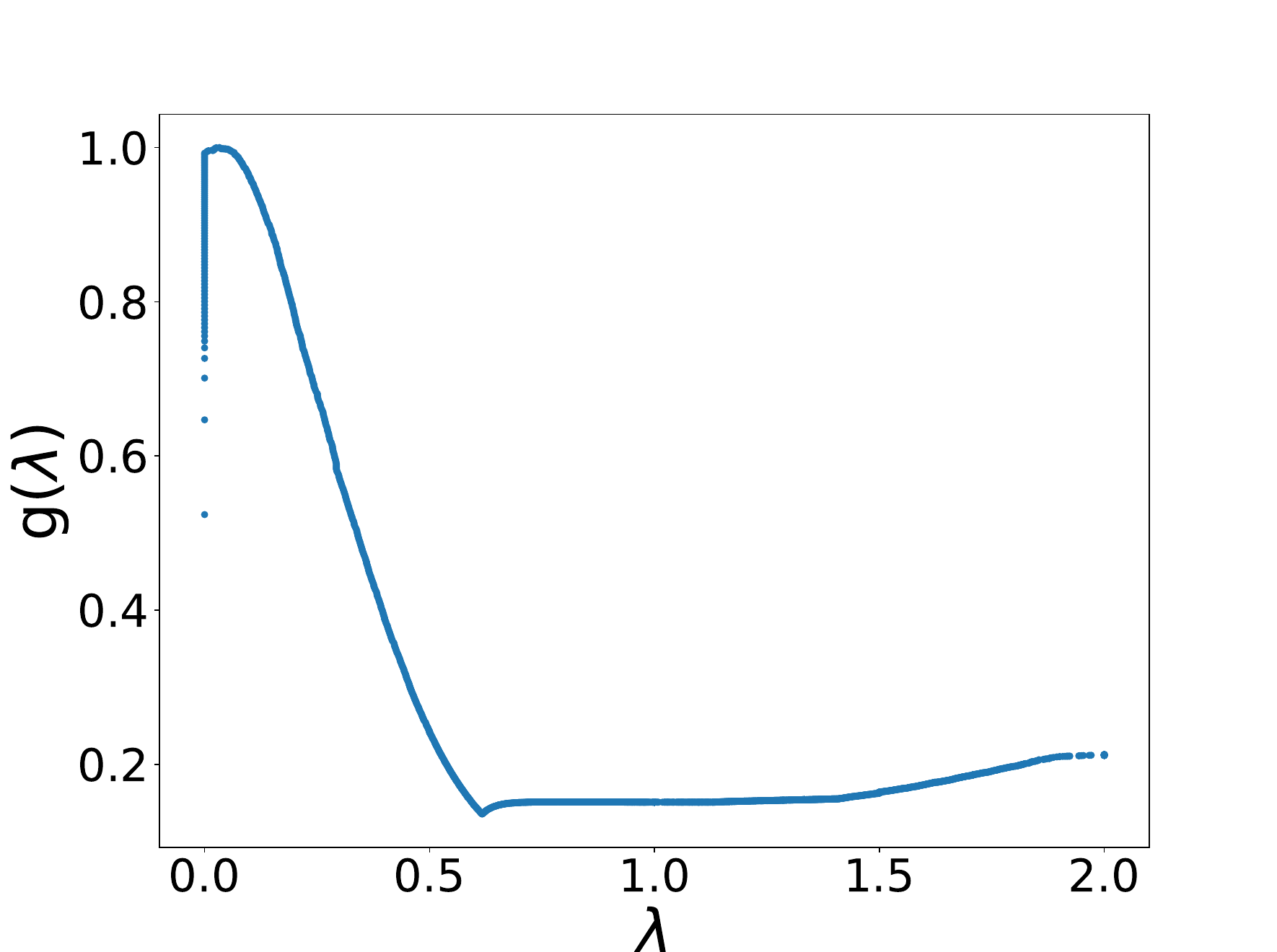}
			\end{minipage}
		}
		\subfigure[CiteSeer]{
			\begin{minipage}[b]{0.28\textwidth}
				\centering
				\includegraphics[width=.85\textwidth]{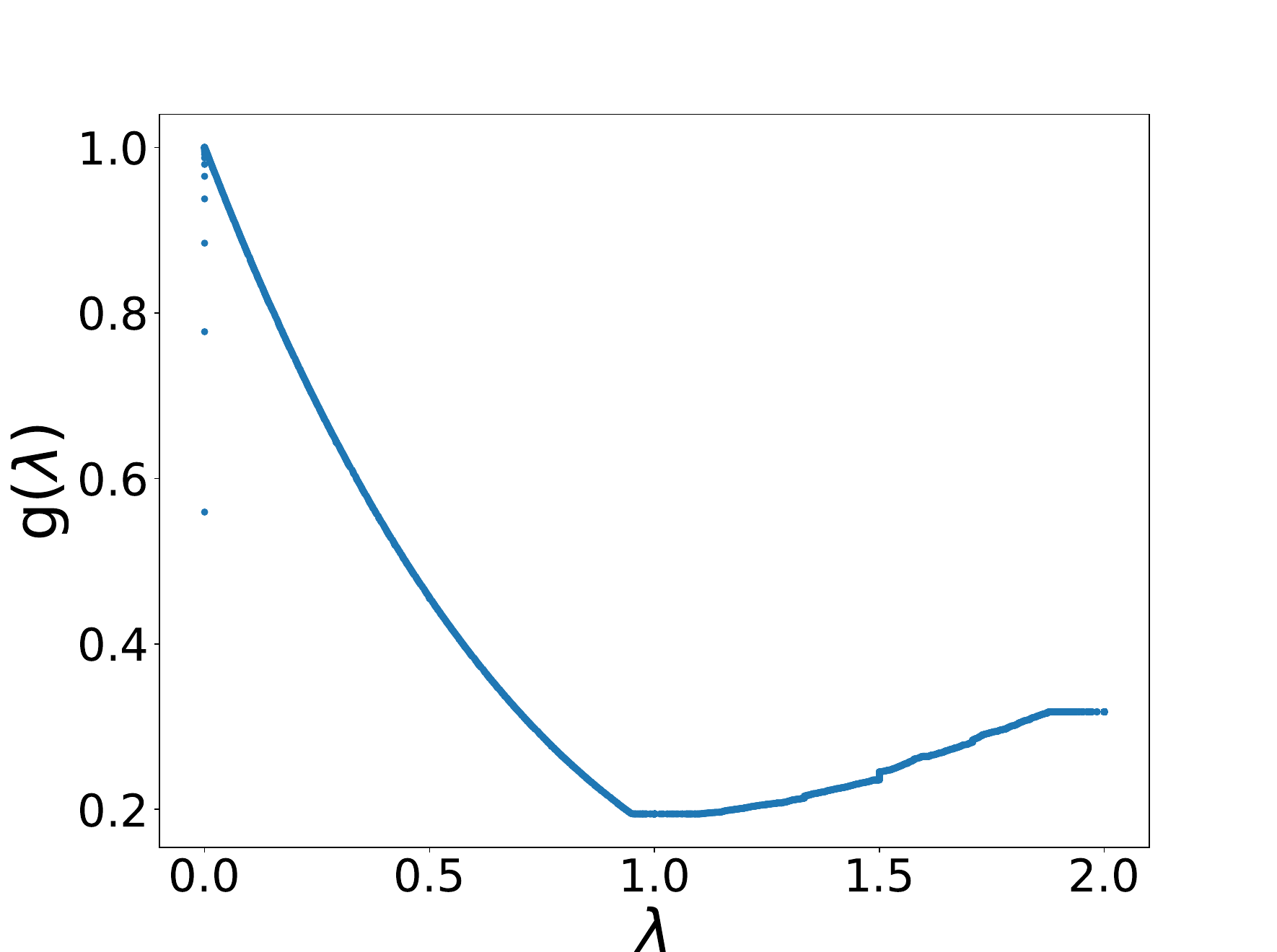}
			\end{minipage}
		}
		\subfigure[Photo]{
			\begin{minipage}[b]{0.28\textwidth}
				\centering
				\includegraphics[width=.85\textwidth]{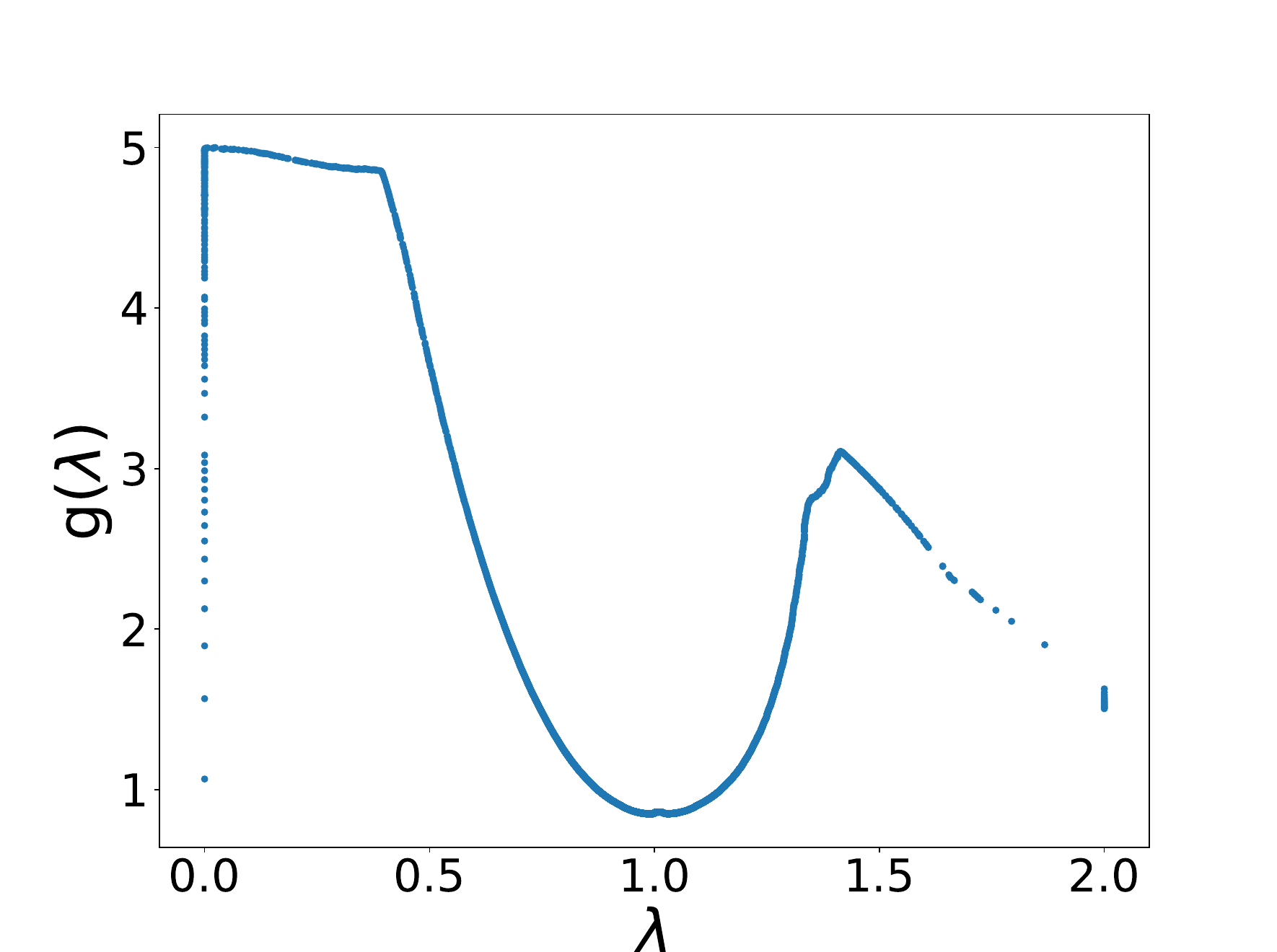}
			\end{minipage}
		}
		\subfigure[Chameleon]{
			\begin{minipage}[b]{0.28\textwidth}
				\centering
				\includegraphics[width=.85\textwidth]{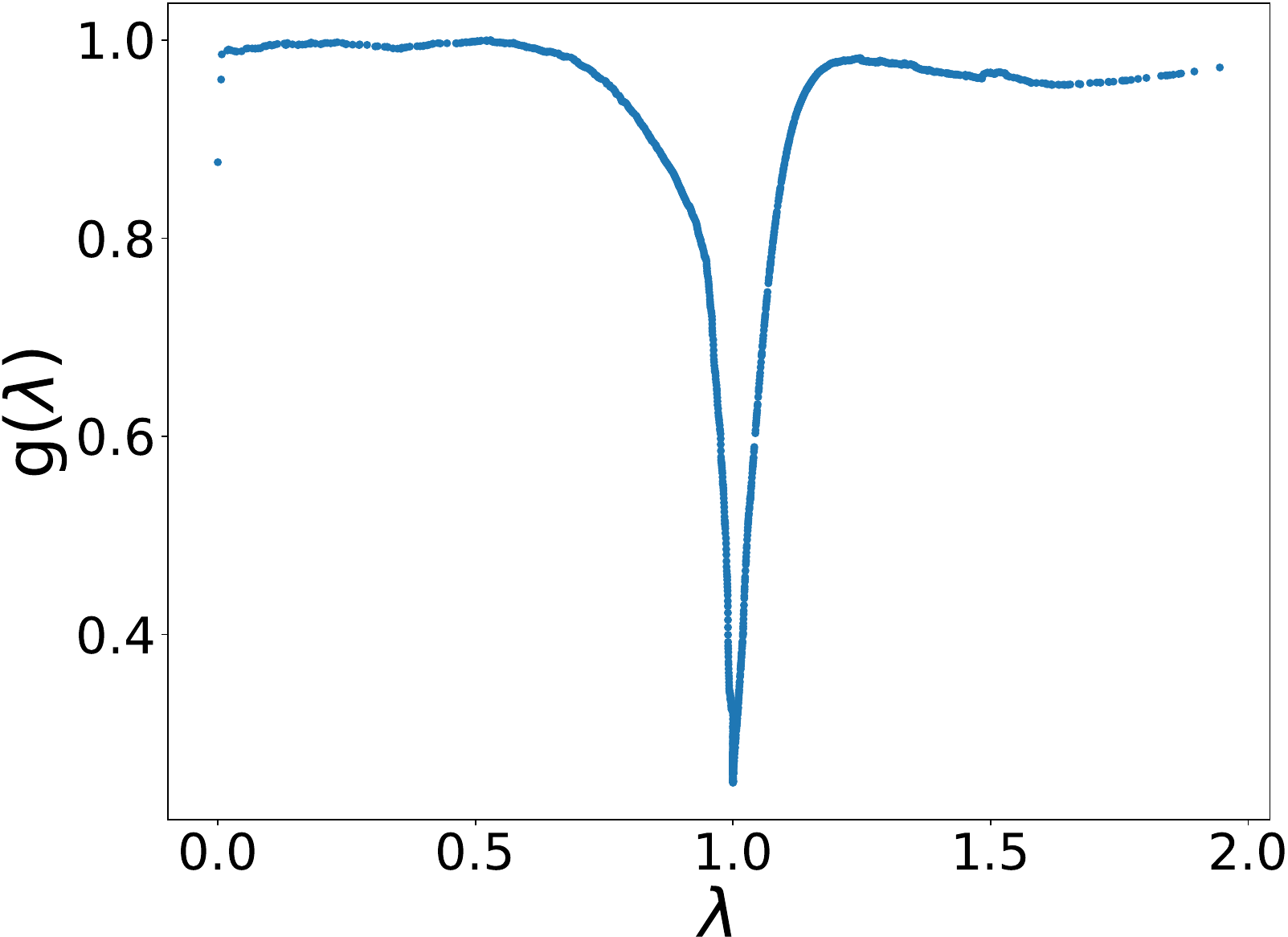}
			\end{minipage}
		}
		\subfigure[Squirrel]{
			\begin{minipage}[b]{0.28\textwidth}
				\centering
				\includegraphics[width=.85\textwidth]{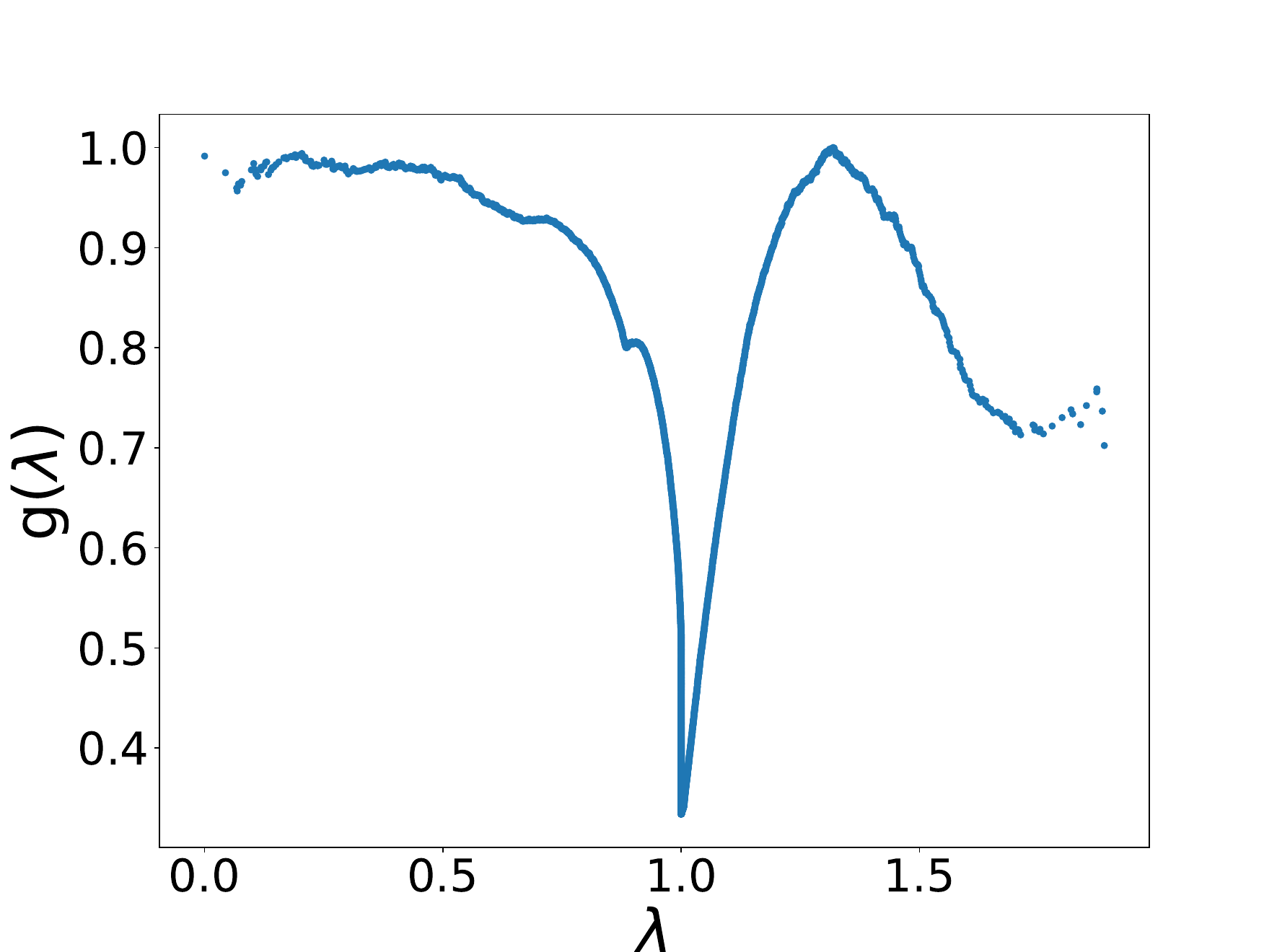}
			\end{minipage}
		}
		\subfigure[Texas]{
			\begin{minipage}[b]{0.28\textwidth}
				\centering
				\includegraphics[width=.85\textwidth]{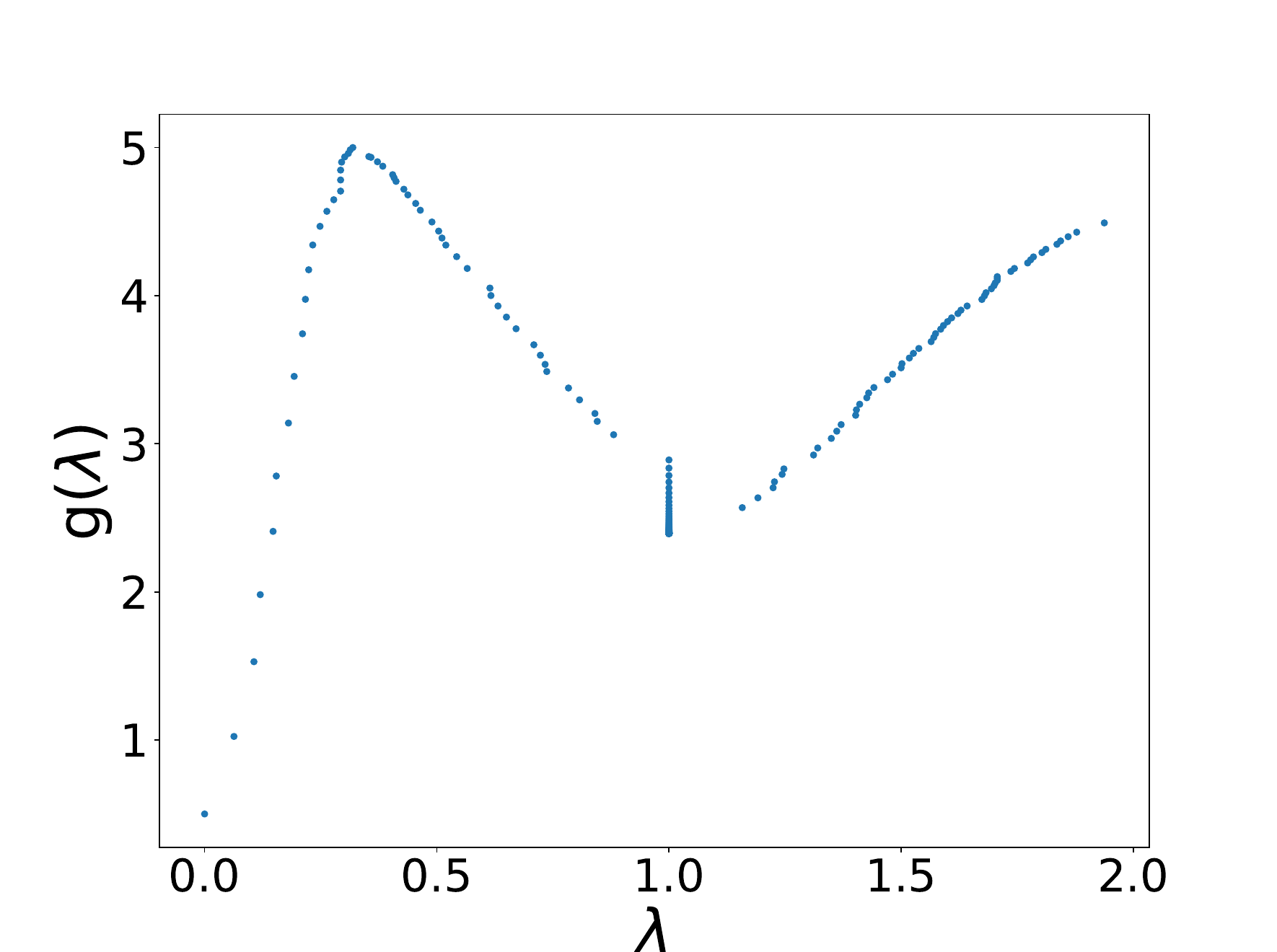}
			\end{minipage}
		}
	\end{center}
	\caption{Illustration of filters learned from real-world datasets by GrassNet.} 
	\label{fig:visulization}
\end{figure}

By visualizing the learned filters from various datasets using GrassNet, we observe that our method adaptively derives different yet appropriate filters for different graphs. Since the SSM graph filter models the discrete frequencies of a graph, GrassNet does not generate a continuous filtering function for the graph dataset. Instead, it derives modulation for each specific frequency, as illustrated in the scatter diagrams shown in Fig.~\ref{fig:visulization}. For homophilic graphs (\textit{e.g.}, Cora, CiteSeer, and Photo), our method derives smooth, low-pass filters, while for heterophilic graphs (\textit{e.g.}, Chameleon, Squirrel, and Texas), it produces more complex, unsmooth, and band-pass filters. In comparison with existing polynomial-based and wavelet-based methods, our method effectively handles challenging scenarios where frequencies are similar or identical (\textit{e.g.}, $\lambda$ equals zero in homophilic graphs and equals one in heterophilic graphs). These results demonstrate that GrassNet excels in managing complex graph spectral signals.

\begin{table}[t!]
	\centering
	\setlength\tabcolsep{6.0pt}
	\renewcommand{\arraystretch}{1.2} 
	\caption{Ablation study results on different datasets with proposed designs.}\label{tab:ablation1}
	\scalebox{.95}{
		\begin{tabular}{ccccc}
			\hline
			& Cora                                                  & CiteSeer                                              & Photo                                                 & Chameleon                                             \\ \hline
			FC            & \begin{tabular}[c]{@{}c@{}}86.33\\ $\pm$0.64\end{tabular} & \begin{tabular}[c]{@{}c@{}}79.90\\ $\pm$0.70\end{tabular} & \begin{tabular}[c]{@{}c@{}}92.05\\ $\pm$1.31\end{tabular} & \begin{tabular}[c]{@{}c@{}}45.93\\ $\pm$1.36\end{tabular} \\ \hline
			RNN           & \begin{tabular}[c]{@{}c@{}}87.75\\ $\pm$0.43\end{tabular} & \begin{tabular}[c]{@{}c@{}}79.82\\ $\pm$0.86\end{tabular} & \begin{tabular}[c]{@{}c@{}}93.77\\ $\pm$0.40\end{tabular} & \begin{tabular}[c]{@{}c@{}}69.91\\ $\pm$1.18\end{tabular} \\ \hline
			LSTM          & \begin{tabular}[c]{@{}c@{}}88.10\\ $\pm$0.56\end{tabular} & \begin{tabular}[c]{@{}c@{}}80.17\\ $\pm$0.71\end{tabular} & \begin{tabular}[c]{@{}c@{}}93.95\\ $\pm$0.18\end{tabular} & \begin{tabular}[c]{@{}c@{}}70.24\\ $\pm$1.55\end{tabular} \\ \hline
			ATTEN         & \begin{tabular}[c]{@{}c@{}}81.61\\ $\pm$3.82\end{tabular} & \begin{tabular}[c]{@{}c@{}}78.40\\ $\pm$1.72\end{tabular} & \begin{tabular}[c]{@{}c@{}}91.33\\ $\pm$3.25\end{tabular} & \begin{tabular}[c]{@{}c@{}}41.18\\ $\pm$1.49\end{tabular} \\ \hline
			SSM-un        & \begin{tabular}[c]{@{}c@{}}\underline{88.24}\\ $\pm$0.56\end{tabular} & \begin{tabular}[c]{@{}c@{}}\underline{81.37}\\ $\pm$0.65\end{tabular} & \begin{tabular}[c]{@{}c@{}}\underline{93.87}\\ $\pm$0.29\end{tabular} & \begin{tabular}[c]{@{}c@{}}\underline{70.96}\\ $\pm$0.99\end{tabular} \\ \hline
			SSM-bi (ours) & \begin{tabular}[c]{@{}c@{}}\textbf{88.59}\\ $\pm$0.69\end{tabular} & \begin{tabular}[c]{@{}c@{}}\textbf{81.51}\\ $\pm$0.67\end{tabular} & \begin{tabular}[c]{@{}c@{}}\textbf{94.49}\\ $\pm$0.27\end{tabular} & \begin{tabular}[c]{@{}c@{}}\textbf{72.71}\\ $\pm$0.98\end{tabular}
			\\ \hline
		\end{tabular}
	}
\end{table}

\begin{figure}[t]
	\setlength{\abovecaptionskip}{.12cm}  
	\begin{center}
		\includegraphics[width=.70\linewidth]{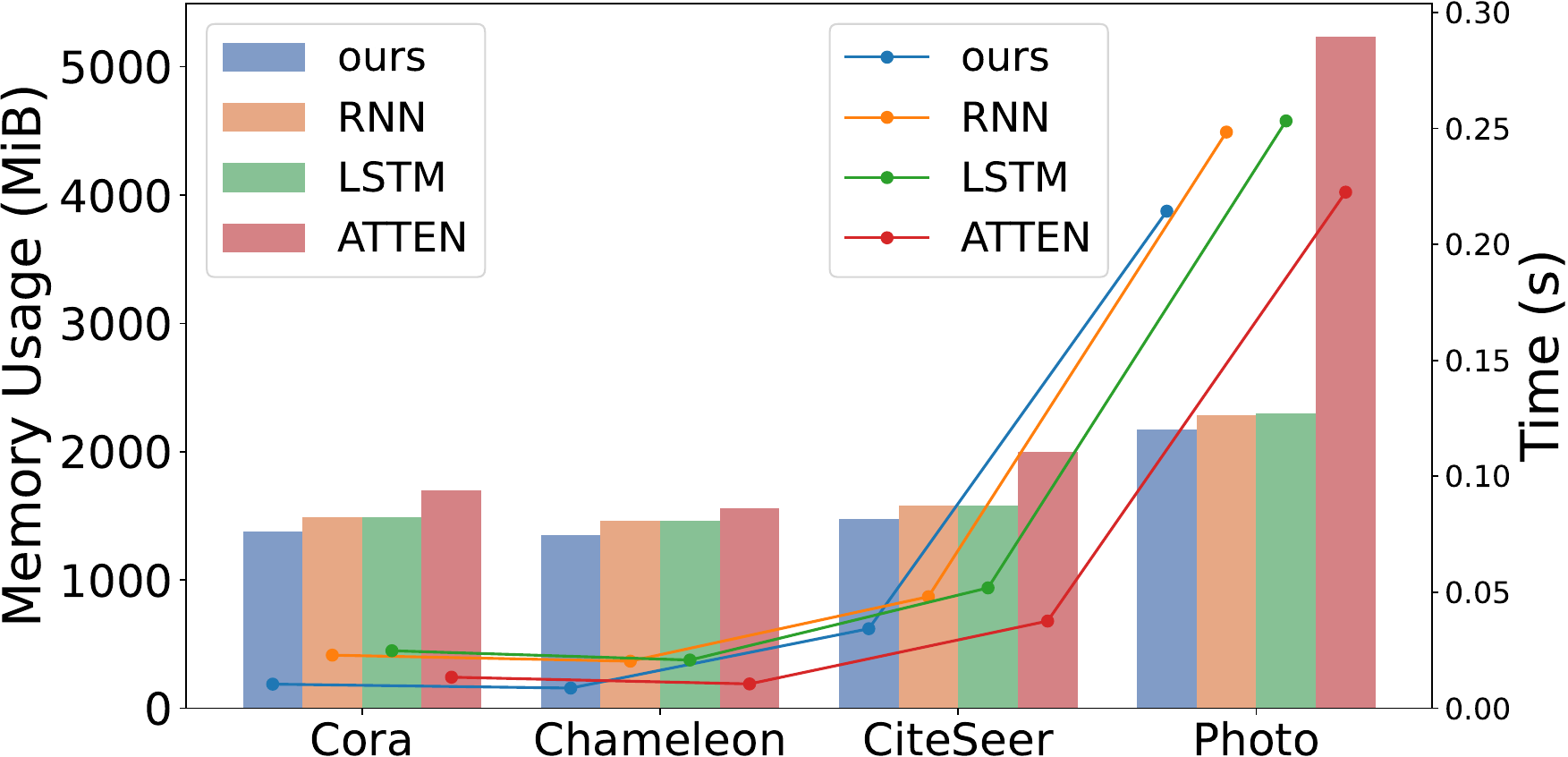}
	\end{center}
	\caption{Memory usage (MiB) and average running time per epoch (s).} 
	\label{fig:time}
\end{figure}

\subsection{Ablation Study}

\subsubsection{Effectiveness of SSM Graph Filter}
We ablate the proposed method to study the importance of designs in GrassNet, focusing particularly on the effectiveness of the SSM graph filter. Five variants are studied and compared with GrassNet. We replace the bi-directional SSM graph filter (\textit{abbr.} SSM-bi) with a fully connected layer (\textit{abbr.} FC) and an undirected SSM graph filter (\textit{abbr.} SSM-un), respectively. Note that for the trial with FC, the spectral filter degrades from the form of Eq.~\ref{eq:poly1} to the form of Eq.~\ref{eq:poly}. We also replace the SSM graph filter with some popular architectures specifically designed for sequence processing, such as recurrent neural network~\cite{sundermeyer2012lstm} (\textit{abbr.} RNN), long short-term memory network~\cite{kim2016long} (\textit{abbr.} LSTM), and attention block~\cite{vaswani2017attention} (\textit{abbr.} ATTEN). The ablation results are included in Tab.~\ref{tab:ablation1}. Comparing FC with other trials, it is evident that the sequential spectral graph filtering paradigm of Eq.~\ref{eq:poly1} contributes significantly to performance. SSM-bi exhibits better results compared with SSM-un, as SSM-bi uses a bi-directional SSM graph filter to fully capture the potential correlations between each frequency, whereas SSM-un only allows each frequency to ``see" those priors to it. Additionally, our method shows overall superior performance compared with RNN, LSTM, and ATTEN, demonstrating the effectiveness of the proposed SSM graph filter.

\subsubsection{Efficiency of SSM Graph Filter}
The comparison of memory usage and average running time per epoch is displayed in Fig.~\ref{fig:time}. Regarding memory usage, ATTEN has the highest space requirement due to the quadratic computational cost associated with the attention mechanism, making it challenging to apply to large-scale graphs. In contrast, RNN, LSTM, and our method benefit from linear space complexity, with the advantage becoming more pronounced as the graph size increases. Our method requires slightly less memory than RNN and LSTM due to its simpler hidden state updates. In terms of time cost, our method is comparable to ATTEN and demonstrates better time efficiency than RNN and LSTM because of the parallel associative scan strategy within SSMs for training parallelizability.

\begin{table}[t!]
	\centering
	\setlength\tabcolsep{6.0pt}
	\renewcommand{\arraystretch}{1.2} 
	\caption{Classification accuracy under two types of perturbed graph topologies.}\label{tab:robustness}
		\begin{tabular}{cccccc}
			\hline
			&         & Cora           & CiteSeer       & Photo          & Chameleon      \\ \hline
			\multirow{4}{*}{rand}  & GCN     & 83.08          & 77.89          & 89.27          & 60.61          \\
			& BernNet & 86.53          & 79.26          & \textbf{90.58} & 64.11          \\
			& WaveNet & 87.35          & 79.67          & 87.25          & 63.68          \\
			& ours    & \textbf{88.18} & \textbf{82.13} & 90.08          & \textbf{65.21} \\  \hline
			\multirow{4}{*}{metis} & GCN     & 79.80          & 74.76          & 87.44          & 55.80          \\
			& BernNet & 85 .22         & 78.58          & 89.27          & 55.36          \\
			& WaveNet & 85.71          & 79.63          & 87.20          & 62.58          \\
			& ours    & \textbf{87.19} & \textbf{81.17} & \textbf{89.73} & \textbf{64.99} \\ \hline
		\end{tabular}
\end{table}

\subsubsection{Robustness of GrassNet}

Existing spectral GNNs exhibit unsatisfactory robustness against edge perturbation, as shown in Fig.~\ref{fig:attack}. To evaluate the robustness of our method, we disrupt the original topology of graph data by removing edges and compare the performance of GrassNet with several representative methods. The experimental settings are consistent with those in Fig.~\ref{fig:attack}, where edges are removed either randomly (\textit{abbr.} rand) or according to the Metis algorithm (\textit{abbr.} metis). The number of removed edges is positively correlated with the original graph structure (1024 for Cora, 907 for CiteSeer, 9022 for Chameleon, and 27704 for Photo). As shown in Tab.~\ref{tab:robustness}, our method demonstrates superior performance under both types of perturbations, particularly with the Metis algorithm that generates more numerically equal frequencies. These results, depicted in Tab.~\ref{tab:robustness}, highlight the effectiveness and robustness of our method.

\section{Conclusion}

In this paper, we delve into the potential of state space models as spectral filters within the context of graph neural networks,  resulting in a novel spectral graph neural network GrassNet. It leverages carefully designed SSM graph filters to effectively model graph spectrum sequences, theoretically offering greater expressive power compared with traditional polynomial and wavelet-based filters. This enhanced capability allows GrassNet to adeptly handle complex spectral patterns within graphs. To the best of our knowledge, this is the first research effort to employ SSMs in the design of spectral filters for GNNs, marking a significant advancement in the field. We conduct extensive experiments on a variety of real-world datasets, which demonstrate the superior effectiveness, efficiency, and robustness of our proposed method.

\newpage

\bibliographystyle{unsrt}  
\bibliography{references}

\newpage
\appendix

\section{APPENDIX}

Here we provide some implementation details of our methods to help readers further understand the algorithms and experiments in this paper.

\section{The Algorithm of GrassNet }

\begin{algorithm}[h!]
	\caption{The GrassNet Algorithm}
	\label{alg:algorithm_ours}
	\textbf{Input}: $G$, $\textbf{X}$, $\textbf{y}_{L}$, \textit{max\_epoch}. \\ 
	\textbf{Output}: $\tilde{\textbf{y}_{U}}$.
	
	\begin{algorithmic}[1] 
		\STATE Compute eigenvalues $\lambda_{i}$ and eigenvector matrix \textbf{U} of $G$;
		\\ \COMMENT{Training process}
		\FOR{\textit{epoch} in [0, 1, ..., \textit{max\_epoch}-1]}
		\STATE Compute the sequence $[\psi(\lambda_{1}), \cdots, \psi(\lambda_{n})]$ by ordering according to the value of $\lambda_{i}$;
		\STATE Compute the sequence of filtering coefficients $[g(\lambda_{1}), \cdots, g(\lambda_{n})]$ using the SSM graph filter as described in Eqs.4-7;
		\STATE Compute the graph convolution output $\tilde{\textbf{X}}$  as described in Eqs.8 and 9;
		\STATE Compute the predictions for the labeled nodes $\tilde{\textbf{y}}_{L}$ using $\tilde{\textbf{X}}_{L}$;
		\STATE Compute the loss according to $\tilde{\textbf{y}}_{L}$ and $\textbf{y}_{L}$;
		\STATE Optimize GrassNet according to the loss;
		\ENDFOR  \\
		\COMMENT{Testing Process}
		\STATE Compute the predictions for the unlabeled nodes $\tilde{\textbf{y}}_{U}$;
		\RETURN $\tilde{\textbf{y}}_{U}$.
	\end{algorithmic}
	
\end{algorithm}

\section{Datasets Statistics}

We evaluate our method on nine benchmark datasets across domains: citation networks (Cora, CiteSeer, PubMed)~\cite{sen2008collective,yang2016revisiting}, Amazon co-purchase graph (Photo)~\cite{mcauley2015image}, Wikipedia graphs (Chameleon, Squirrel)~\cite{rozemberczki2021multi}, actor co-occurrence graph (Actor)~\cite{pei2019geom} and webpage graphs from WebKB (Texas,  Cornell)~\cite{pei2019geom}. The datasets adopted are representative which describe diverse real-life scenarios. Some of them are highly homophilic while others are heterophilic. The detailed statistics of the datasets are summarized in Tab.~\ref{tab:dataset}.

\begin{table}[h]
	\centering
	\setlength\tabcolsep{6.pt}
	\renewcommand{\arraystretch}{1.3} 
	\caption{Datasets statistics.}\label{tab:dataset}
	\begin{tabular}{lcccc}
		\hline
		& Nodes & Edges  & Features & Classes \\ \hline
		Cora      & 2708  & 5278   & 1433     & 7       \\
		CiteSeer  & 3327  & 4552   & 3703     & 6       \\
		PubMed    & 19717 & 44324  & 500      & 5       \\
		Photo     & 7650  & 119081 & 745      & 8       \\
		Chameleon & 2277  & 31371  & 2325     & 5       \\
		Squirrel  & 5201  & 198353 & 2089     & 5       \\
		Actor     & 7600  & 26659  & 932      & 5       \\
		Texas     & 183   & 279    & 1703     & 5       \\
		Cornell   & 183   & 277    & 1703     & 5      \\ \hline
	\end{tabular}
\end{table}

\begin{itemize}
	\item \textbf{Cora}, \textbf{CiteSeer} and \textbf{PubMed}~\cite{sen2008collective} are three classic homophilic citation networks. In these networks, nodes correspond to academic papers, and edges signify the citation links between papers. The node features are derived from bag-of-word representations of the papers, and the labels categorize each paper into specific research topics.
	
	\item \textbf{Photo}~\cite{mcauley2015image} is segments of the Amazon co-purchase graph, where nodes denotes goods, edges indicate two goods are frequently bought together, node features are bag-of-words encoded product reviews, and class labels are given by the product category.
	
	\item \textbf{Chameleon} and \textbf{Squirrel}~\cite{rozemberczki2021multi} are two heterophilic networks derived from Wikipedia. In these networks, nodes represent Wikipedia web pages, and edges correspond to hyperlinks between these pages. The features are comprised of informative nouns extracted from the Wikipedia content, while the labels reflect the average traffic of each web page.

	\item \textbf{Actor}~\cite{pei2019geom} is a heterophilic actor co-occurrence network where nodes represent actors, and edges signify that two actors have appeared together in the same movie. The features are derived from keywords found on the actors' Wikipedia pages, while the labels consist of significant words associated with each actor. 
	
	\item \textbf{Cornell} and \textbf{Texas}~\cite{pei2019geom} are heterophilic networks from the WebKB1 project representing computer science departments at three universities. Nodes are departmental web pages, edges represent hyperlinks, features are derived using bag-of-words, and labels categorize page types. These networks illustrate heterophilic connections where linked pages often differ in type.

\end{itemize}

\section{Experimental Settings}

We conduct the semi-supervised node classification task across all baseline methods, where the node set is randomly divided into training, validation, and test sets with a 60\%/20\%/20\% split. To ensure fairness, we generate 10 random splits using different seeds and evaluate all approaches on these identical splits, reporting the average performance for each method. Our method is compared against ten baseline GNN methods, including MLP~\cite{taud2018multilayer}, GCN~\cite{kipf2016semi}, GAT~\cite{velivckovic2018graph}, ChebNet~\cite{defferrard2016convolutional}, SplineCNN~\cite{fey2018splinecnn}, APPNP~\cite{gasteiger2018predict}, GPR-GNN~\cite{chien2021adaptive}, and the recently proposed BernNet~\cite{he2021bernnet}, ChebNetII~\cite{he2022convolutional}, and WaveNet~\cite{yang2024wavenet}. For the training of our GrassNet, we use the Adam optimization algorithm with an appropriate L2 penalty specific to Adam. The evaluation metric used is accuracy (acc), presented with a 95\% confidence interval.

The GrassNet code is built on the GNNs in the PyTorch version of the Deep Graph Library (DGL)\cite{wang2019deep}. We search for the optimal scaling factor $\gamma$ within \{0.01, 0.1, 0.5, 1, 5, 10\}, the number of hidden units in our model within \{8, 16, 32, 64, 128\}, and the number of fully connected layers in Eq.8 within \{1, 2, 3, 4, 5, 6\}. For training, we search for the learning rate within {0.001, 0.01, 0.1} and weight decay within \{0.0005, 0\}. All experiments are conducted on an AMD EPYC 7542 32-Core Processor with an Nvidia GeForce RTX 3090. The hyper-parameters used in our model are listed in Tab.\ref{tab:hyper}.

\begin{table}[h!]
	\setlength\tabcolsep{2.5pt}
	\renewcommand{\arraystretch}{1.3} 
	\caption{Hyper-parameters of proposed method on real-world datasets.}
	\begin{center}
		\begin{tabular}{lcccccc}
			\hline
			\multicolumn{1}{c}{} & $\gamma$ & \begin{tabular}[c]{@{}c@{}}Learning\\ rate\end{tabular} & \begin{tabular}[c]{@{}c@{}}Hidden\\ units\end{tabular} & \begin{tabular}[c]{@{}c@{}}FC\\ layers\end{tabular} & \begin{tabular}[c]{@{}c@{}}Weight\\ decay\end{tabular} & \begin{tabular}[c]{@{}c@{}}Training\\ epoch\end{tabular} \\ \hline
			Cora                 & 1     & 0.1                                                     & 16                                                     & 1                                                   & 0.0005                                                 & 1000                                                     \\
			CiteSeer             & 1     & 0.1                                                     & 16                                                     & 1                                                   & 0.0005                                                 & 1000                                                     \\
			PubMed               & 5     & 0.1                                                     & 32                                                     & 1                                                   & 0.0005                                                 & 1000                                                     \\
			Photo                & 5     & 0.01                                                    & 32                                                     & 1                                                   & 0.0005                                                 & 1000                                                     \\
			Chameleon            & 1     & 0.01                                                    & 16                                                     & 2                                                   & 0.0005                                                 & 1000                                                     \\
			Squirrel             & 1     & 0.01                                                    & 32                                                     & 2                                                   & 0.0005                                                 & 1000                                                     \\
			Actor                & 0.1   & 0.01                                                    & 16                                                     & 1                                                   & 0.0005                                                 & 1000                                                     \\
			Texas                & 5     & 0.1                                                     & 8                                                      & 1                                                   & 0.0005                                                 & 1000                                                     \\
			Cornell              & 0.5   & 0.1                                                     & 8                                                      & 1                                                   & 0.0005                                                 & 1000                        \\ \hline                            
		\end{tabular}
	\end{center}
	\label{tab:hyper}
\end{table}

\newpage
\section{Codes of baseline models}\label{app:codes}

We implemented all baseline models using the official code and leveraged libraries such as PyTorch Geometric~\cite{fey2019fast}, DGL~\cite{wang2019deep}. The URLs of all the baseline approaches we used are listed in Tab.~\ref{tab:url}. Our code will be made public upon the acceptance of this paper.

\begin{table}[t!]
	\setlength\tabcolsep{5.2pt}
	\renewcommand{\arraystretch}{1.5} 
	\caption{Codes of baseline approaches.}
	\begin{center}
		\begin{tabular}{lc}
			\hline
			& URL                                                  \\  \hline
			GCN        & \url{https://github.com/tkipf/gcn}                         \\
			APPNP      & \url{https://github.com/benedekrozemberczki/APPNP}         \\
			GAT        & \url{https://github.com/PetarV-/GAT}                      \\
			ChebNet    & \url{https://github.com/dsgiitr/graph_nets}   \\
			GPR-GNN    &   \url{https://github.com/jianhao2016/GPRGNN}  \\
			SplineCNN  &   \url{https://github.com/LONG-9621/SplineCNN}  \\
			BernNet    &    \url{https://github.com/ivam-he/BernNet}    \\
			ChebNetII  &  \url{https://github.com/ivam-he/ChebNetII}   \\
			WaveNet     &   \url{https://github.com/Bufordyang/WaveNet}  \\ \hline
		\end{tabular}
	\end{center}
	\label{tab:url}
\end{table}

\end{document}